\documentclass[letterpaper]{article} 
\usepackage{aaai24}  
\usepackage{times}  
\usepackage{helvet}  
\usepackage{courier}  
\usepackage[hyphens]{url}  
\usepackage{graphicx} 
\usepackage{natbib}  
\usepackage{caption} 
\usepackage{amsthm}
\newtheorem{Problem}{Problem}
\newtheorem{Theorem}{Theorem}
\newtheorem{Proposition}{Proposition}
\newtheorem{Lemma}{Lemma}

\newtheorem{Remark}{Remark}
\newtheorem{Definition}{Definition}

\usepackage{hyphenat}
\usepackage{microtype}
\usepackage{booktabs} 
\usepackage{bbold}
\usepackage{subfigure}
\usepackage{enumitem}
\usepackage{relsize}
\usepackage{dsfont}
\usepackage{multirow}
\usepackage{threeparttable}
\usepackage{stmaryrd}
\usepackage{longtable}
\usepackage{algorithm}
\usepackage{algorithmic}

\usepackage{amsmath,amsfonts,bm}

\def\eqref#1{equation~\ref{#1}}

\def\1{\bm{1}}

\def\vm{{\bm{m}}}

\def\evm{{m}}

\def\mM{{\mathbf{M}}}

\DeclareMathAlphabet{\mathsfit}{\encodingdefault}{\sfdefault}{m}{sl}
\SetMathAlphabet{\mathsfit}{bold}{\encodingdefault}{\sfdefault}{bx}{n}

\def\gE{{\mathcal{E}}}

\def\gV{{\mathcal{V}}}

\def\sG{{\mathbb{G}}}

\def\sR{{\mathbb{R}}}

\def\sV{{\mathbb{V}}}

\def\emA{{A}}

\DeclareMathOperator*{\argmax}{arg\,max}
\DeclareMathOperator*{\argmin}{arg\,min}

\usepackage{newfloat}
\usepackage{listings}

\usepackage{newfloat}
\usepackage{listings}
\DeclareCaptionStyle{ruled}{labelfont=normalfont,labelsep=colon,strut=off} 
\floatstyle{ruled}
\newfloat{listing}{tb}{lst}{}
\floatname{listing}{Listing}
\usepackage{url}
\usepackage{graphicx}
\usepackage{subfigure}
\usepackage{amssymb}
\usepackage{algorithm}
\usepackage{algorithmic}
\newcommand{\bashapes}{BA-Shapes}
\newcommand{\bacom}{BA-Community}
\newcommand{\treec}{Tree-Circles}
\newcommand{\treeg}{Tree-Grid}
\newcommand{\bamo}{BA-2motifs}
\newcommand{\bafour}{BA-4motifs}
\newcommand{\mutag}{MUTAG}
\newcommand{\ours}{K-FactExplainer }
\usepackage{amsmath}

\usepackage{array}

\title{Factorized Explainer for Graph Neural Networks}
\author{
    Rundong Huang\textsuperscript{\rm 1},
    Farhad Shirani\textsuperscript{\rm 2}$^*$,
    Dongsheng Luo\textsuperscript{\rm 2}\thanks{Corresponding authors.}
}
\affiliations{
    \textsuperscript{\rm 1}Technical University of Munich, Munich, Germany\\
    \textsuperscript{\rm 2}Florida International University, Miami, U.S.\\
    rundong.huang@tum.de, \{fshirani,dluo\}@fiu.edu
}

\usepackage{bibentry}

\begin{document}

\maketitle

\begin{abstract}
Graph Neural Networks (GNNs) have received increasing attention due to their ability to learn from graph-structured data. To open the black-box of these deep learning models, post-hoc instance-level explanation methods have been proposed to understand GNN predictions. These methods seek to discover substructures that explain the prediction behavior of a trained GNN.  In this paper, we show analytically that for a large class of explanation tasks, conventional approaches, which are based on the principle of graph information bottleneck (GIB), admit trivial solutions that do not align with the notion of explainability. Instead, we argue that a modified GIB principle may be used to avoid the aforementioned trivial solutions. We further introduce a novel factorized explanation model with theoretical performance guarantees. The modified GIB is used to analyze the structural properties of the proposed factorized explainer. We conduct extensive experiments on both synthetic and real-world datasets to validate the effectiveness of our proposed factorized explainer.
\end{abstract}
\section{Introduction}
Graph-structured data is ubiquitous in real-world applications, manifesting in various domains such as social networks~\cite{Fan2019GraphNN}, molecular structures~\cite{Mansimov2019MolecularGP,Chereda2019UtilizingMN}, and knowledge graphs~\cite{liu2022federated}. This has led to significant interest in learning methodologies specific to graphical data, particularly, graph neural networks (GNNs). GNNs commonly employ message-passing mechanisms, recursively transmitting and fusing messages among neighboring nodes on graphs. Thus, the learned node representation captures both node attributes and neighborhood information, thereby enabling diverse downstream tasks such as node classification~\cite{kipf2017semisupervised,veličković2018graph}, graph classification~\cite{xu2018how}, and link prediction~\cite{lu2022graph}.

Despite the success of GNNs in a wide range of domains, their inherent  ``black-box'' nature and lack of interpretability, a characteristic shared among many contemporary machine learning methods,  is a major roadblock in their utility in sensitive application scenarios such as autonomous decision systems. To address this, various GNN explanation methods have been proposed to understand the graph-structured data and associated deep graph learning models~\cite{ying2019gnnexplainer,luo2020parameterized,yuan2022explainability}.  In particular, post-hoc instance-level explanation methods provide an effective way to identify determinant substructures in the input graph, which plays a vital role in trustworthy deployments~\cite{ying2019gnnexplainer,luo2020parameterized}. In the context of graph classification, the objective of graph explanation methods is, given a graph $G$, to extract a \textit{minimal} and \textit{sufficient} subgraph, $G^*$, that can be used to determine the instance label, $Y$. The  Graph Information Bottleneck principle (GIB)~\cite{wu2020graph} provides an intuitive principle that is widely adopted as a practical instantiation. At a high level, the GIB principle finds the subgraph $G^*$ which minimizes the mutual information between the original graph $G$ and the subgraph $G^*$ and maximizes the mutual information between the subgraph $G^*$ and instance label $Y$ by minimizing $I(G, G^*)-\alpha I(G^*, Y)$, where the hyperparameter $\alpha>0$ captures the tradeoff between minimality and informativeness of $G^*$ ~\cite{miao2022interpretable}. As an example, the GNNExplainer method operates by finding a learnable edge mask matrix, which is optimized by the GIB objective~\cite{ying2019gnnexplainer}.  The PGExplainer also uses a GIB-based objective and incorporates a parametric generator to learn explanatory subgraphs from the model's output~\cite{luo2020parameterized}.

There are several limitations in the existing explainability approaches. First, as shown analytically in this work, existing GIB-based methods suffer from perceptually unrealistic explanations. Specifically, we show that in a wide-range of statistical scenarios, the original GIB formulation of the explainability problem has a trivial solution where the achieved explanation $G^*$ \textit{signals} the predicted value of $Y$, but is independent of the input graph $G$, otherwise. That is, the Markov chain $G^*\leftrightarrow Y \leftrightarrow G$ holds. As a result, the explanation $G^*$ optimizing the GIB objective may consist of a few disconnected edges and fails to align with the high-level notion of explainability.  To alleviate this problem, PGExplainer includes an ad-hoc connectivity constraint as the regularization term~\cite{luo2020parameterized}. However, without theoretical guarantees, the effectiveness of the extra regularization is marginal in more complicated datasets~\cite{shan2021reinforcement}.  Second, although previous parametric explanation methods, such as PGExplainer~\cite{luo2020parameterized} and ReFine~\cite{wang2021towards}, are efficient in the inductive setting, these methods neglect the existence of multiple motifs, which is routinely observed in real-life datasets. For example, In the MUTAG dataset~\cite{debnath1991structure}, both chemical components $NO_2$ and $NH_2$, which can be considered as explanation subgraphs, contribute to the positive mutagenicity. Existing methods over-simplify the relationship between motifs and labels to one-to-one, leading to inaccuracy in real-life applications. 

To address these issues,  we first analytically investigate the pitfalls of the application of the GIB principle in explanation tasks from an information theoretic perspective, and propose a modified GIB principle that avoids these issues. To further improve the inductive performance, we propose a new framework to unify existing parametric methods and show that their suboptimality is caused by their locality property and the lossy aggregation step in GNNs. We further propose a straightforward and effective factorization-based explanation method to break the limitation of existing local explanation functions.  We summarize our main contributions as follows. 
\begin{itemize}
    \item For the first time, we point out that the gap between the practical objective function (GIB) and high-level objective is non-negligible in the most popular post-hoc explanation framework for graph neural networks. 
    \item We derive a generalized framework to unify existing parametric explanation methods and theoretically analyze their pitfalls in achieving accurate explanations in complicated real-life datasets.  We further propose a straightforward explanation method with a solid theoretical foundation to achieve better generalization capacity. 
    \item Comprehensive empirical studies on both synthetic and real-life datasets demonstrate that our method can consistently improve the quality of the explanations.
\end{itemize}

\section{Preliminary}
\subsection{Notations and Problem Definition}
A graph $G$ is parameterized by a quadruple $(\mathcal{V}, \mathcal{E}; \mathbf{Z}, \mathbf{A})$, where  i) $\mathcal{V} = \{v_1, v_2, ..., v_n\}$ is the node/vertex\footnote{We use node and vertex interchangeably.} set, ii) $ \mathcal{E} \subseteq \mathcal{V} \times \mathcal{V}$ is the edge set, iii) $ \mathbf{Z} \in \sR^{n\times d}$ is the feature matrix, where the $i$th row of $\mathbf{Z}$, denoted by $ \mathbf{z}_i  \in \sR^{1\times d} $, is the $d$-dimensional feature vector associated with node $v_i, i\in [n]$, and iv) the adjacency matrix $ \mathbf{A} \in \{0,1\}^{n\times n}$ is determined by the edge set $\mathcal{E}$, where $\emA_{ij} = 1$ if $(v_i,v_j)\in \mathcal{E}$, $\emA_{ij} = 0$, otherwise. We write $|G|$ and $|\mathcal{E}|$ interchangeably to denote the number of edges of $G$.

For graph classification task, each graph $G_i$ has a label $Y_i \in \mathcal{C}$, with a GNN model $f$ trained to classify $G_i$ into its class,   i.e., {$f:G\mapsto \{1,2,\cdots,|\mathcal{C}|\}$}. 
For the node classification task, each graph $G_i$ denotes a $K$-hop sub-graph centered around node $v_i$, with a GNN model $f$ trained to predict the label for node $v_i$ based on the node representation of $v_i$ learned from $G_i$.

\begin{Problem} [Post-hoc Instance-level GNN Explanation~\cite{yuan2022explainability,luo2020parameterized,ying2019gnnexplainer}]
\label{prob:exp}
Given a trained GNN model $f$, for an arbitrary input graph $G= (\mathcal{V}, \mathcal{E}; \mathbf{Z}, \mathbf{A})$, the goal of post-hoc instance-level GNN explanation is to find a subgraph $G^{*}=\Psi(G)$ that can `explain'\footnote{The notion of explainability is made precise in the sequel, where the modified GIB principle is introduced.} the prediction of $f$ on $G$. The mapping $\Psi:G\mapsto G^*$ is called the explanation function. 
\end{Problem}
Informative feature selection has been well studied in non-graph structured data~\cite{li2017feature},  and traditional methods, such as concrete autoencoder~\cite{abid2019concrete}, can be directly extended to explain features in GNNs. In this paper, we focus on discovering important typologies. Formally, the obtained explanation $G^{*}$ is characterized by a binary mask $\mM \in \{0, 1\}^{n\times n}$ on the adjacency matrix, e.g., $G^{*} = ( \gV, \gE, \mathbf{A}\odot \mathbf{M}; \mathbf{Z})$, where $\odot$ is elements-wise multiplication. The mask highlights components of $G$ which influence the output of $f$. 

\subsection{Graph Information Bottleneck}
The GIB principle refers to the graphical version of the Information Bottleneck (IB) principle~\cite{tishby2015deep} which offers an intuitive measure for learning dense representations. It is based on the notion that an optimal representation should contain \textit{minimal} and \textit{sufficient} information for the downstream prediction task. Recently, a high-level unification of several existing post hoc GNN explanation methods, such as GNNExplainer~\cite{ying2019gnnexplainer}, and PGExplainer~\cite{luo2020parameterized} was provided using this concept ~\cite{wu2020graph,miao2022interpretable,yu2021graph}.
Formally, prior works have represented the objective of finding an explanation graph $G^*$ in $G$ as follows: 
\begin{equation}
    \label{eq:gib}
    G^*\triangleq \argmin_{P_{G'|G}: \mathbb{E}(|G'|)\leq \gamma} I(G, G')-\alpha I(G',Y),
\end{equation}
where $G^*$ is the explanation subgraph, $\gamma\in \mathbb{N}$ is the maximum expected size (number of edges) of the explanation, 
$Y$ is the original or ground truth label, and $\alpha$ is a hyper-parameter capturing the trade-off between minimality and sufficiency constraints.
At a high level, the 
GIB formulation given in \eqref{eq:gib} selects the minimal explanation $G'$, by minimizing $I(G,G')$ and imposing $\mathbb{E}(|G'|)\leq \gamma$, that inherits only the most indicative information from $G$ to predict the label $Y$, by maximizing $I(G',Y)$, while avoiding imposing potentially biased constraints, such the connectivity of the selected subgraphs and exact maximum size constraints~\cite{miao2022interpretable}. Note that from the definition of mutual information, we have $I(G', Y) = H(Y)-H(Y|G')$, where the entropy $H(Y)$ is static and independent of the explanation process. Thus, minimizing the mutual information between the explanation subgraph $G'$ and $Y$ can be reformulated as maximizing the conditional entropy of $Y$ given $G'$. That is:

\begin{equation}
    \label{eq:gibp}
    G^*= \argmin_{P_{G'|G}: \mathbb{E}(|G'|)\leq \gamma} {I(G, G')+\alpha H(Y|G')}.
\end{equation}

\section{Graph Information Bottleneck for Explanation}
\label{sec:gub}
In this section, we study several pitfalls arising from the application of the GIB principle to explanation tasks. We demonstrate that, for a broad range of learning tasks, the original GIB formulation of the explainability problem has a trivial solution that does not align with the intuitive notion of explainability. We propose a modified version of the GIB principle that avoids this trivial solution and is applicable in constructing GNN explanation methods. The analytical derivations in subsequent sections will focus on this modified GIB principle. To elaborate, we argue that the optimization given in \eqref{eq:gibp} is prone to \textit{signaling issues} and, in general, does not fully align theoretically with the notion of explainability. More precisely, the GIB formulation allows for an explanation algorithm to output $G^*$ which \textit{signals} the predicted value of $Y$, but is independent of the input graph $G$ otherwise. To state this more concretely, we consider the class of statistically degraded classification tasks defined in the following.

\begin{Definition}[\textbf{Statistically Degraded Classification}]
\label{def:1}
Consider a classification task characterized by the triple $(\mathcal{X},\mathcal{Y},P_{\mathbf{X},Y})$, where $\mathcal{X}$ represents the feature space, $\mathcal{Y}$ denotes the set of output labels, and $P_{\mathbf{X},Y}$ characterizes the joint distribution of features and labels. The classification task is called statistically degraded\footnote{Statistical degradedness has its origins in the field of information theory, particularly in communication and estimation applications \cite{el2010lecture}} if there exists a function $h:\mathcal{X}\to \mathcal{Y}$ such that the Markov chain $\mathbf{X}\leftrightarrow h(\mathbf{X})\leftrightarrow Y$ holds. That is, $h(\mathbf{X})$ is a sufficient statistic for $\mathbf{X}$ w.r.t. $Y$.
\end{Definition}

\begin{Remark}
Any deterministic classification task, where there exists a function $h:\mathcal{X}\to \mathcal{Y}$ such that $h(\mathbf{X})=Y$, is statistically degraded.
\end{Remark}

\begin{Remark}
There are classification tasks that are not statistically degraded. For instance, let us consider a classification task in which the feature vector is $\mathbf{X} = (X_1, X_2)$, where $X_1$ and $X_2$ are independent binary symmetric variables. Let the label $Y$ be equal to $X_1$ with probability $p \in (0,1)$ and equal to $X_2$, otherwise. By exhaustively searching over all 16 possible choices of $h(\mathbf{X})$, it can be verified that no Boolean function $h(\mathbf{X})$ exists such that the relationship $\mathbf{X} \leftrightarrow h(\mathbf{X}) \leftrightarrow Y$ holds. Consequently, the classification task $(\mathcal{X}, \mathcal{Y}, P_{\mathbf{X}, Y})$ is not statistically degraded.
\end{Remark}

\begin{Remark}
    Note that for the statistically degraded task defined in Definition \ref{def:1}, the optimal classifier $f^*(\mathbf{X})$ is equal to the sufficient statistic $h(\mathbf{X})$. 
\end{Remark}

 In order to show the limitations of the GIB in fully encapsulating the concept of explainability, in the sequel we focus on statistically degraded classification tasks involving graph inputs. That is, we take $\mathbf{X}=G$, where $G$ is the input graph. 
The next lemma shows that, for any statistically degraded task, there exists an explanation function $\Psi(\cdot)$ which optimizes the GIB objective function (\eqref{eq:gibp}), and whose output is independent of $G$ given $h(\cdot)$. That is, although the explanation algorithm is optimal in the GIB sense, it does not provide any additional information about the input of the classifier, in addition to the information that the classifier output label $h(G)$ readily provides. 
\begin{Theorem}
\label{th:1}
Consider a statistically degraded graph classification task, parametrized by $(P_{G,Y},h(\cdot))$, where $P_{G,Y}$ is the joint distribution of input graphs and their labels,  and $h: \mathcal{G}\to \mathcal{Y}$ is such that $G\leftrightarrow h(G) \leftrightarrow Y$ holds. For any $\alpha>0$, there exists an explanation algorithm $\Psi_{\alpha}(\cdot)$ such that $G'\triangleq \Psi_{\alpha}(G)$ optimizes the objective function in \eqref{eq:gibp} and $\Psi_{\alpha}(G) \leftrightarrow h(G) \leftrightarrow G$  holds.
\end{Theorem}

The proof relies on the following modified data processing inequality. 

\begin{Lemma}[\textbf{Modified Data Processing Inequality}]
\label{lem:1}
    Let $A,B$ and $C$ be random variables satisfying the Markov chain $A\leftrightarrow B\leftrightarrow C$. Define the random variable $A'$ such that $P_{A'|C}= P_{A|C}$ and $A,B \leftrightarrow C \leftrightarrow A'$. Then, 
    \begin{align*}
        I(A',B)\leq I(A,B).
    \end{align*}
\end{Lemma}
The proof of Lemma \ref{lem:1} and Theorem \ref{th:1} are included in Appendix.

As shown by Theorem \ref{lem:1}, the original GIB formulation does not fully align with the notion of explainability. Consequently, we adopt the following modified objective function:
\begin{equation}
    \label{eq:gibp-ce}
   G^*\triangleq  \argmin_{P_{G'|G}:\mathbb{E}(|G'|)\leq \gamma} {I(G, G')+\alpha CE(Y,Y')},
\end{equation}
where $Y'\triangleq f(G')$ is the predicted label of $G'$ made by the model to be explained $f$, and the cross-entropy $\text{CE}(Y,Y')$ between the ground truth label $Y$ and $Y'$ is used in place of $H(Y|G')$ in the original GIB. The modified GIB avoids the signaling issues in Theorem \ref{lem:1}, by comparing the correct label $Y$ with the prediction output $Y'$ based on the original model $f(\cdot)$. This is in contrast with the original GIB principle which measures the mutual information $I(Y,G')$, which provides a general measure of how well $Y$ can be predicted from $G'$ (via Fano's inequality \cite{el2010lecture}), without relating this prediction to the original model $f(\cdot)$. It should be mentioned that several recent works have also adopted this modified GIB formulation~\cite{ying2019gnnexplainer,luo2020parameterized}. However, the rationale provided in these earlier studies was that the modified GIB serves as a computationally efficient approximation for the original GIB, rather than addressing the limitations of the original GIB shown in Theorem \ref{th:1}.

\section{K-FactExplainer for Graph Neural Networks}
\label{sec:method}
In this section, we first theoretically show that existing parametric explainers based on the GIB objective, such as PGExplainer~\cite{luo2020parameterized}, are subject to two sources of inaccuracies: locality and lossy aggregation. Then we propose a straightforward and effective approach to mitigating the problem. In the subsequent sections, we provide simulation results that corroborate these theoretical predictions.

\subsection{Theoretical Analysis}
We first define the general class of \textit{local explanation methods}.

\begin{Definition}[\textbf{Geodisc Restricted Graph}]
    Given a graph $G=(\mathcal{V},\mathcal{E};\mathbf{Z},\mathbf{A})$, node $v\in \mathcal{V}$, and a radius $r \in \mathbb{N}$, the $(v,r)$-restriction of $G$ is the graph $G_{v,r}= (\mathcal{V}_{v,r},\mathcal{E}_{v,r};\mathbf{Z}_{v,r},\mathbf{A}_{v,r})$, where 
    \begin{itemize}
        \item $\mathcal{V}_{v,r}\triangleq \{v'| d(v,v')\leq r\}$, where $d(\cdot,\cdot)$ is the geodisc distance. 
        \item $\mathcal{E}_{v,r}\triangleq \{(v_i,v_j)| e\in \mathcal{E}, v_i,v_j\in \mathcal{V}_{v,r}\}$. 
        \item $\mathbf{Z}_{v,r}$ consists of feature vectors in $\mathbf{Z}$ corresponding to $v\in \mathcal{V}_{v,r}$.
        \item $\mathbf{A}_{v,r}$ is the adjacency matrix corresponding to $\mathcal{E}_{v,r}$. 
    \end{itemize}
\end{Definition}

\begin{Definition}[\textbf{Local Explanation Methods}]
    Consider a graph classification task $(\mathcal{G},\mathcal{Y}, P_{G,Y})$, a classification function $f:\mathcal{G}\to \mathcal{Y}$, a parameter $r\in \mathbb{N}$, and an explanation function $\Psi:\mathcal{G}\to \mathcal{G}$, where $\mathcal{G}$ is the set of all possible input graphs, and $\mathcal{Y}$ is the set of output labels.
    Let $G'=\Psi(G)= (\mathcal{V}',\mathcal{E}';\mathbf{Z}',\mathbf{A}')$. The explanation function $\Psi(\cdot)$ is called an $r$-local explanation function if:
    \begin{enumerate}
        \item The Markov chain $\mathbb{1}(v\in \mathcal{V}') \leftrightarrow G_{v,r} \leftrightarrow G$ holds for all $v\in \mathcal{V}$, where $\mathbb{1}(\cdot)$ is the indicator function.
        \item The edge $(v,v')$ is in $\mathcal{E}'$ if and only if $v,v'\in \mathcal{V}'$ and $e\in \mathcal{E}$. 
    \end{enumerate}
    \label{def:loc}
\end{Definition}
The first condition in Definition \ref{def:loc} requires that the presence of each vertex $v$ in the explanation $G'$ only depends on its neighboring vertices in $G$ which are within its $r$ local neighborhood. The second condition requires that $G'$ be a subgraph of $G$. It is straightforward to show that various explanation methods such as PGExplainer are local explanation methods due to the boundedness of their corresponding computation graphs. This is formalized in the following proposition. 
\begin{Proposition}[\textbf{Locality of PGExplainer}]
    Consider a graph classification task $(\mathcal{G},\mathcal{Y}, P_{G,Y})$ and an $\ell$ layer GNN classifier $f(\cdot)$, for some $\ell\in \mathbb{N}$. Then, any explanation $\Psi(\cdot)$ for $f(\cdot)$ produced using the PGExplainer is an $\ell$-local explanation function.  
\end{Proposition}

Next, we argue that local explanation methods cannot be optimal in the modified GIB sense for various classification tasks. Furthermore, we argue that this issue may be mitigated by the addition of a hyperparameter $k$ as described in subsequent sections in the context of the K-FactExplainer. 

To provide concrete analytical arguments, we focus on a specific graph classification task, where the class labels are binary, the input graph has binary-valued edges, and the output label is a function of a set of indicator motifs. To elaborate, we assume that the label to be predicted is $Y=\max\{E_1,E_2,\cdots,E_s\}$, where $E_i, i\in [s]$ are Bernoulli variables, and if $E_i=1$, then $g_i\subseteq G$ for some fixed subgraphs $g_i, i\in [s]$. In the explainability literature, each of the subgraphs $g_i, i\in [s]$  is called a motif for label $Y=1$. Let us define $G_e= \bigcup_{i\in [s]}g_i\mathbb{1}(E_i=1)$. So that $G_e$ is the union of all the edges in the motifs that are present in $G$, and it is empty if $Y=0$. Formally, the classification task under consideration is characterized by the following joint distribution:
\begin{equation}
\begin{aligned}
    &P_{G,Y}(g,y) \\
    =& \sum_{e^s,g_0}P_{E^s}(e^s)P_{G_0}(g_0)\mathbb{1}(y=\max_{i\in [s]} e_i,g=g_0\cup g_e),
    \label{eq:2}
\end{aligned}    
\end{equation}
where $e^s\in \{0,1\}^s$, $g_e\triangleq  \bigcup_{i\in [s]}g_i\mathbb{1}(e_i=1)$, and $G_0$ is the ``irrelevant'' edges in $G$ with respect to the label $Y$.
    
\begin{Remark}
    The graph classification task on the MUTAG dataset is an instance of the above classification scenario, where there are two motifs, corresponding to the existence of $NH_2$ and $NO_2$ chemical groups, respectively \cite{ying2018graph}. Similarly, the BA-4Motif classification task considered in the Appendix can be posed in the form of \eqref{eq:2}.
\end{Remark}
In graph classification tasks characterized by \eqref{eq:2}, if the label of $G$ is one, then at least one of the motifs is present in $G$. Note that the reverse may not be true as the motifs may randomly appear in the `irrelevant' graph $G_0$ due to its probabilistic nature. A natural choice for the explanation function $\Psi(\cdot)$ of a classifier  $f(\cdot)$ for this task is one which outputs one of the motifs present in $G$ if $f(G)=1$. For instance, in the MUTAG classification task, an explainer should output $NH_2$ or $NO_2$ subgraphs if the output label is equal to one. In the following, we argue that,  in classification tasks involving more than one motif local explanation methods cannot produce the motifs accurately. Hence their output does not align with the natural explanation outlined above and is not optimal in the modified GIB sense.
To make the result concrete, we further make the following simplifying assumptions: 
    \\i) The graph $G_0$ is Erd\"os-R\'enyi with parameter $p\in (0,\frac{1}{2})$: 
    \[P_{G_0}(g_0)= p^{|g_0|}(1-p)^{\frac{n(n-1)}{2}-|g_0|}.\]
    \\ii) There exists $r,r'>0$ such that the geodisc radius and geodisc diameter of $g_i$ are less than or equal to $r$ and $r'$, respectively, for all $i\in [s]$. 
    \\iii) The geodisc distance between $g_i$ and $g_j$ is greater than $r$ for all $i\neq j$.
    \\iv) $E_i, i\in [s]$ are jointly independent Bernoulli variables with parameter $p_i$, where $P_{G_0}(g_i)\leq p_i$. 

\begin{Theorem}[\textbf{Suboptimality of Local Explanation Functions}]
Let $r,r'\in \mathbb{N}$. For the graph classification task described in \eqref{eq:2}, the following hold: 
\\a) The optimal Bayes classification rule $f^*(g)$ is equal to $\mathbb{1}(\exists i\in [s]: g_i\subseteq g)$. \\b) For any r-local explanation function, there exists $\alpha'>0$ such that the explanation is suboptimal for $f^*$ in the modified GIB sense for all $\alpha>\alpha'$ and $\gamma$ equal to maximum number of edges of $g_i, i\in [s]$.   
\\c) There exists an integer $k\leq s$, a parameter $\alpha'>0$, a collection of $r'$-local explanation functions $\Psi_i(\cdot), i\in [k]$, and an explanation function $\Psi^*$, such that for all inputs $g$, we have $\Psi(g)\in \{\Psi_1(g),\Psi_2(g),\cdots,\Psi_{k}(g)\}$ and $\Psi^*$ is optimal in the modified GIB sense for all $\alpha>\alpha'$ and $\gamma$ equal to maximum number of edges of $g_i, i\in [s]$.   
\label{th:2}
\end{Theorem}

The proof of Theorem \ref{th:2} is provided in the Appendix.

\begin{figure*}[t!]
	\centering
	\includegraphics[width= \textwidth,draft=false]{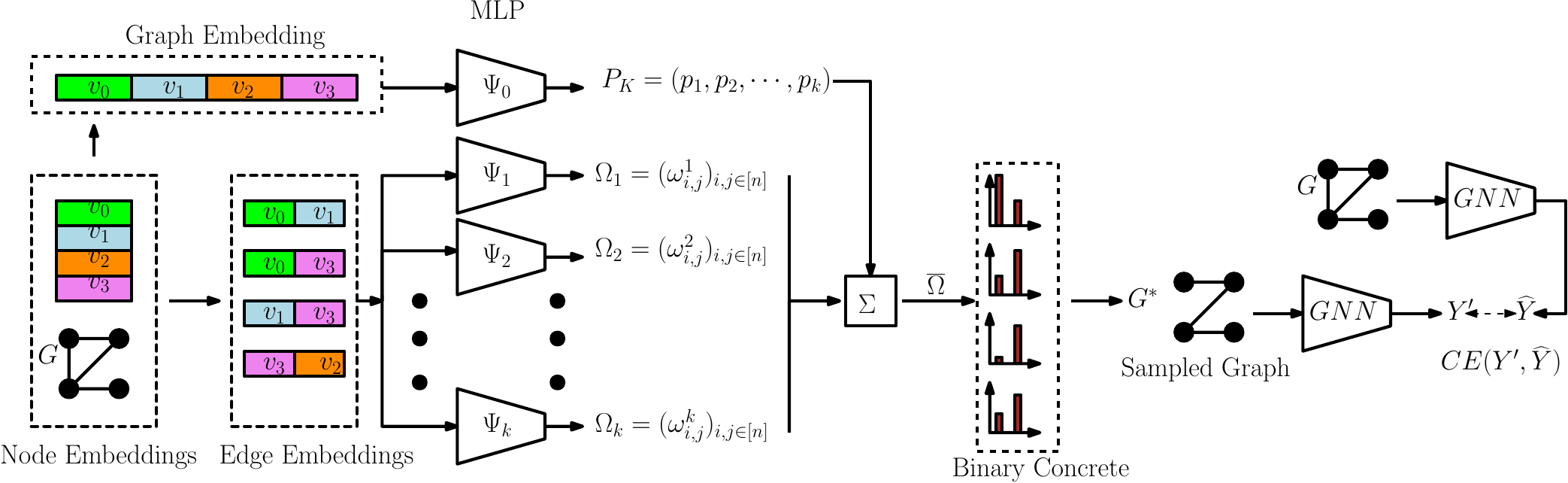}
	\caption{ Illustration of K-FactExplainer method. Starting from the left, the node embeddings for graph G are produced using the original GNN. The edge embeddings and graph embedding are produced by concatenating the node embeddings. The MLP $\Psi_0$ assigns weight to the outputs of PGExplainer MLPs $\Psi_{t}, t\in[k]$. The resulting vector of edge probabilities $\overline{\Omega}\triangleq (\sum_{t=1}^kp_t\omega^t_{i,j})_{i,j\in [n]}$ is used to produce the sampled explanation graph $G^*$. The explanation is fed to the original GNN and the output label is compared with the original prediction. The training proceeds by minimizing the cross-entropy term $CE(Y',\widehat{Y})$, where $\widehat{Y}$ is GNN prediction for the original input graph $G$.} 
	\label{fig:Ov}
\end{figure*}

 Theorem \ref{th:2} can be interpreted as follows: for graph classification tasks with more than one motif, although local explanation methods are not optimal in general, one can ``patch'' together several local explanation methods $\Psi_1(\cdot),\Psi_2(\cdot),\cdots,\Psi_k(\cdot)$ into an explanation method $\Psi^*(\cdot)$, such that i) for any given input $g$, the output of  $\Psi^*(g)$ is equal to the output of one of the explanation functions $\Psi_1(g),\Psi_2(g),\cdots,\Psi_k(g)$, and  ii) $\Psi^*(\cdot)$ is optimal in the modified GIB sense. This insight motivates the K-FactExplainer method introduced in the following section. 

\begin{Remark}
\label{rem:loss}
     Theorem \ref{th:2} implies that local explanation methods are not optimal in multi-motif classification tasks.  It should be noted that even in single-motif tasks, post-hoc methods which rely on GNN generated node embeddings for explanations would perform suboptimally. The reason is that the aggregator function which is used to generate the embeddings is lossy (is not a one-to-one function) and potentially loses information during the GNN aggregation step. This can also be observed in the simulation results provided in the sequel, where we apply our proposed K-FactExplainer method and show gains compared to the state of the art in both multi-motif and single-motif scenarios.  
\end{Remark}

\subsection{K-FactExplainer and a Bootstrapping Algorithm}
\label{sec:met}
Motivated by the insights gained by the analytical results in the previous section, we propose a new graph explanation method. An overview of the proposed method is shown in Figure \ref{fig:Ov}. To describe the method, let  $f(\cdot)$ be the GNN which we wish to explain. Let $\mathbf{Z}_i, i\in [n]$ denote the node embedding for node $v_i, i\in [n]$ produced by $f(\cdot)$. We construct the edge embeddings $\mathbf{Z}_{i,j}=(\mathbf{Z}_i,\mathbf{Z}_j), i,j\in [n]$ and graph embedding $\mathbf{Z}=(\mathbf{Z}_i, i\in [n])$ by concatenating the edge embeddings. Let $k\in \mathbb{N}$ be the upper-bound on the number of necessary local explainers from Theorem \ref{th:2}. 
We consider $k$ multi-layer neural networks (MLPs) denoted by $\Psi_t, t\in [k]$. Each MLP $\Psi_t$ individually operates in a similar fashion as the MLP used in the PGExplainer method. That is, $\Psi_t$ operates on each edge embedding $(\mathbf{Z}_i,\mathbf{Z}_j)$ individually, and outputs a Bernoulli parameter $\omega_{i,j}^t\in [0,1]$.
The parameter $\omega^t_{i,j}\in [0,1]$ can be viewed as the probability that the edge $(v_i,v_j)$ is in the sampled explanation graph.  Based on the insights provided by Theorem \ref{th:2}, we wish to patch together the outputs of $\Psi_t, t\in [k]$ to overcome the locality issue in explaining GNNs in multi-motif classification tasks. This is achieved by including the additional MLP $\Psi_0$ which takes the graph embedding $\mathbf{Z}$ as input and outputs the probability distribution $P_K$ on the alphabet $[k]$. At a high level, the MLP $\Psi_0$ provides a global view of the input graph, whereas each of the $\Psi_t, t\in [k]$ MLPs provide a local perspective of the input graph. 
The outputs $(\omega^t_{i,j})_{i,j\in [n]}$ of $\Psi_t$ are linearly combined with weights associated with $P_K(t), t\in [k]$ and the resulting vector of Bernoulli probabilities $\overline{\Omega}=(\sum_{t=1}^kP_K(t)\omega^t_{i,j})_{i,j\in [n]}$ is used to sample the edges of the input graph $G$ and produce the explanation graph $G^*$. In the training phase, $G^*$ is fed to $f(\cdot)$ to produce the prediction $Y'$. Training is performed by minimizing the cross-entropy term $CE(Y',\widehat{Y})$, where $\widehat{Y}=f(G)$ is the label prediction of the GNN given input $G$. The next proposition provides an algorithm to bound the value of $k$, which determines the number of MLPs which need to be trained.

\begin{table*}[ht]
\centering
\scalebox{0.85}{
  \begin{tabular}{c|cccccc}
    \hline
 & \bashapes & \bacom & \treec& \treeg &\bamo &\mutag \\
    \hline
     GRAD    &0.882      &0.750   &0.905  &0.667&0.717 &0.783\\ 
    ATT     &0.815      &0.739   &0.824  &0.612&0.674 &0.765\\
    RGExp.  & 0.985$_{\pm0.013}$            & 0.919$_{\pm0.017}$            & 0.787$_{\pm0.099}$    & \textbf{0.927}$_{\pm0.032}$    & 0.657$_{\pm0.107}$    & 0.873$_{\pm0.028}$\\
    DEGREE  & 0.993$_{\pm0.005}$            & 0.957$_{\pm0.010}$            & \textbf{0.902}$_{\pm0.022}$    & 0.925$_{\pm0.040}$    & 0.755$_{\pm0.135}$    & 0.773$_{\pm0.029}$\\
    \hline
    GNNExp. & 0.742$_{\pm0.006}$            & 0.708$_{\pm0.004}$            & 0.540$_{\pm0.017}$    & 0.714$_{\pm0.002}$    & 0.499$_{\pm0.004}$    & 0.606$_{\pm0.003}$\\
    PGExp.  & 0.999$_{\pm0.000}$   & 0.825$_{\pm0.040}$            & 0.760$_{\pm0.014}$    & 0.679$_{\pm0.008}$    & 0.566$_{\pm0.004}$    & 0.843$_{\pm0.162}$ \\
    \hline
    \ours    & \textbf{1.000}$_{\pm0.000}$   & \textbf{0.974}$_{\pm0.004}$   & 0.779$_{\pm0.004}$           & 0.770$_{\pm0.004}$           & \textbf{0.821}$_{\pm0.005}$& \textbf{0.915}$_{\pm0.010}$ \\ 
    \hline
  \end{tabular}
  }
    \caption{Explanation faithfulness in terms of AUC-ROC on edges under six datasets. The higher, the better. Our approach achieves consistent improvements over GIB-based explanation methods.} 
    \label{tab:baselines} 
\end{table*}

\begin{Definition}[\textbf{Minimal r-Cover}]
    Given a random graph $G=(\mathcal{V},\mathcal{E};\mathbf{Z},\mathbf{A})$ and non-negative integer $r$, the collection $\mathsf{P}=(\mathcal{P}_1,\mathcal{P}_2,\cdots, \mathcal{P}_{|\mathsf{P}|}), \mathcal{P}_j \subseteq \mathcal{V}$ is called
    an $r$-cover of $G$ if 
    \\i) Each partition element $\mathcal{P}_j$ has geodisc diameter at most equal to $r$, and
    \\ii) $P(\mathcal{V}^*\subseteq \cup_{j\in [|\mathsf{P}|]}\mathcal{P}_{j})=1$, where $\mathcal{V}^*$ denotes the set of vertices of $G$ which are not isolated.
    \\The cover is called minimal if $r$ is the smallest integer for which an r-cover of $G$ exists.
\end{Definition}

\begin{Proposition}[\textbf{Bounding the Number of MLPs}]
Consider the setup in Theorem \ref{th:2}. The parameter $k$, the number of $r'$-local explainers needed to achieve optimal GIB performance, can be upper-bounded by $k^*$ if $G_e$ has a minimal $r'$-cover with $k^*$ elements.
\\Particularly, if the classifier to be explained, $f(\cdot)$, is a GNN with $\ell$ layers, and $\ell$ is greater than or equal to the largest geodisc diameter of the motifs $g_i, i\in [s]$, then $k$ can be upper-bounded by $s$.
\label{prop:2}
\end{Proposition}
The proof of Proposition \ref{prop:2} is provided in the Appendix.

Proposition \ref{prop:2} provides a method to find an upper-bound on $k$; however, it requires that the motifs 
 be known beforehand, so that $G_e$ is known and the size of its minimal cover can be computed. In practice, we do not know the motifs before the start of the explanation process, since the explanation task would be trivial otherwise. We provide an approximate solution, where instead of finding the minimal cover for $G_e$, we use a bootstrapping method in which we find the minimal cover for the explanation graphs produced by another pre-trained explainer, e.g., a PGExplainer. To elaborate,  It takes the GNN model to be explained $f$, a set of training input graphs $\sG$, and a post-hoc explainer $\Psi$ as input. In our simulations, we adopt PGExplainer as the post-hoc explainer $\Psi$. Other explanation methods such as GNNExplainer~\cite{ying2019gnnexplainer} can also be used in this step.  For each graph $G\in \sG$, we first apply the explainer $\Psi$ on $G$ to get the initial explanation graph, whose nodes are listed in $\sV_e$ and edge mask matrix is denoted by $\mM$. This is used as an estimate for $G_e$. To find its minimal cover, we rank the nodes in $\sV_e$ based on their degrees and initialize $k'=0$. For each step, we select a node $v$ from $\sV_e$ and extract its $k^*$-hop neighborhood graph, $G^{(l)}_v$.  Then, we remove all nodes in  $G^{(l)}_v$ from $\sV_e$. After that, we add a count to $k'$ and select the next node in $\sV_e$ until $|\sV_e|=0$. We iterate all graphs in $\sG$ and report the maximum value of $k'$ as $\hat{k}$, the estimate of $k$.  A detailed algorithm can be found in Appendix.

\section{Related Work}
Graph neural networks (GNNs) have gained increasing attention in recent years due to the need for analyzing graph data structures~\cite{kipf2017semisupervised,veličković2018graph, xu2018how, feng2020graph,satorras2021n,bouritsas2022improving}. In general, GNNs model messages from node representations and then propagate messages with message-passing mechanisms to update representations. GNNs have been successfully applied in various graph mining tasks, such as node classification~\cite{kipf2017semisupervised}, link prediction~\cite{zhang2018link}, and graph classification~\cite{xu2018how}. Despite their popularity, akin to other deep learning methodologies, GNNs operate as black box models, which means their functioning can be hard to comprehend, even when the message passing techniques and parameters used are known. Furthermore, GNNs stand apart from conventional deep neural networks that assume instances are identically and independently distributed. GNNs instead integrate node features with graph topology, which complicates the interpretability issue.

Recent studies have aimed to interpret GNN models and offer explanations for their predictions~\cite{ying2019gnnexplainer,luo2020parameterized,yuan2020xgnn,yuan2022explainability,yuan2021explainability, lin2021generative, wang2022gnninterpreter,miao2023interpretable,fang2023cooperative,ma2022clear,zhang2023mixupexplainer}. These methods generally fall into two categories based on granularity: i) instance-level explanation~\cite{ying2019gnnexplainer,zhang2022gstarx}, which explains predictions for each instance by identifying significant substructures; and ii) model-level explanation~\cite{yuan2020xgnn,wang2022gnninterpreter,azzolin2023global}, designed to understand global decision rules incorporated by the target GNN. Among these methods, Post-hoc explanation ones~\cite{ying2019gnnexplainer,luo2020parameterized,yuan2021explainability}, which employ another model or approach to explain a target GNN.  Post-hoc explanations have the advantage of being model-agnostic, meaning they can be applied to a variety of GNNs. Therefore, our work focuses on post-hoc instance-level explanations~\cite{ying2019gnnexplainer}, that is, identifying crucial instance-wise substructures for each input to explain the prediction using a trained GNN model. A detailed survey can be found in ~\cite{yuan2022explainability}.
\begin{table*}[h!]
\centering
\scalebox{0.9}{
  \begin{tabular}{c|cccccc}
    \hline
 & \bashapes & \bacom & \treec& \treeg &\bamo &\mutag \\
    \hline
    PGExp. & 0.999$_{\pm0.000}$& 0.825$_{\pm0.040}$& 0.760$_{\pm0.014}$& 0.679$_{\pm0.008}$& 0.566$_{\pm0.004}$& 0.843$_{\pm0.162}$ \\
    $k=1$  & \textbf{1.000}$_{\pm0.000}$& 0.850$_{\pm0.047}$& 0.758$_{\pm0.023}$& 0.711$_{\pm0.011}$& \underline{0.580}$_{\pm0.041}$& 0.769$_{\pm0.119}$\\
    $k=2$  & \underline{\textbf{1.000}}$_{\pm0.000}$ & 0.880$_{\pm0.023}$& \textbf{0.779}$_{\pm0.018}$& \underline{0.707}$_{\pm0.570}$& 0.581$_{\pm0.039}$& 0.801$_{\pm0.105}$\\   
    $k=3$  & \textbf{1.000}$_{\pm0.000}$& \underline{0.902}$_{\pm0.022}$& \underline{0.772}$_{\pm0.012}$& 0.710$_{\pm0.005}$& 0.586$_{\pm0.034}$& 0.895$_{\pm0.034}$\\
    $k=5$  & \textbf{1.000}$_{\pm0.000}$ & 0.899$_{\pm0.011}$& 0.768$_{\pm0.013}$& 0.709$_{\pm0.006}$& 0.573$_{\pm0.044}$& 0.892$_{\pm0.030}$\\
    $k=10$ & \textbf{1.000}$_{\pm0.000}$& 0.926$_{\pm0.012}$& 0.774$_{\pm0.006}$& 0.706$_{\pm0.004}$& 0.578$_{\pm0.039}$& \textbf{0.915}$_{\pm0.021}$\\
    $k=20$ & \textbf{1.000}$_{\pm0.000}$& 0.938$_{\pm0.013}$& 0.778$_{\pm0.006}$& 0.704$_{\pm0.002}$& 0.586$_{\pm0.032}$& 0.911$_{\pm0.014}$\\
    $k=60$ & \textbf{1.000}$_{\pm0.000}$& \textbf{0.952}$_{\pm0.011}$& 0.778$_{\pm0.004}$& \textbf{0.770}$_{\pm0.004}$& \textbf{0.588}$_{\pm0.030}$& \underline{\textbf{0.915}}$_{\pm0.010}$\\
\hline
  \end{tabular}
  }
      \caption{Explanation performances w.r.t. $k$. We use underlines to denote $k$ selected by the proposed method.}
    \label{tab:tuningK} 
\end{table*}

\section{Experimental Study}
In this section, we empirically verify the effectiveness and efficiency of the proposed \ours by explaining both node and graph classifications. We also conduct extensive studies to verify our theoretical claims. Due to the space limitation, detailed experimental setups, full experimental results, and extensive experiments are presented in Appendix.

\begin{table*}[ht]
\centering
\scalebox{0.9}{
  \begin{tabular}{c|c c c c c c c c  }
  \hline
 & $k=1$& $k=2$ & $k=3$ &$k=5$&$k=10$&$k=20$&$k=60$&\\
\hline
BA-Community(20)&0.850&0.880&0.902&0.899&0.926&0.938&0.952&\\
BA-Community(80)&0.893&0.899&0.895&0.894&0.895&0.895&0.897&\\
\hline
Tree-Circles(20) & 0.758& 0.779& 0.772& 0.768& 0.774& 0.778& 0.778&\\
Tree-Circles(80) & 0.871& 0.871& 0.871& 0.870& 0.871& 0.871& 0.870&\\
\hline
\end{tabular}
}
\caption{Effects of lossy aggregation on GNNs with different hidden layer sizes}
\label{tab:lossy}
\end{table*}

\subsubsection{Experimental Setup.} 
We compare our method with representative GIB-based explanation methods, GNNExplainer~\cite{ying2019gnnexplainer} and PGExplainer~\cite{luo2020parameterized}, classic explanation methods,  GRAD~\cite{ying2019gnnexplainer} and ATT~\cite{veličković2018graph}, and SOTA methods, RG-Explainer~\cite{shan2021reinforcement} and DEGREE ~\cite{feng2022degree}.  We follow the routinely adopted framework to set up our experiments~\cite{ying2019gnnexplainer,luo2020parameterized}. Six benchmark datasets with ground truth explanations are used for evaluation, with \bashapes, \bacom, \treec, and \treeg~\cite{ying2019gnnexplainer} for the node classification task, and \bamo~\cite{luo2020parameterized} and MUTAG~\cite{debnath1991structure} for the graph classification task. For each dataset, we train a graph neural network model to perform the node or graph classification task. Each model is a three-layer GNN with a hidden size of 20, followed by an MLP that maps these embeddings to the number of classes. After training the model, we apply the \ours and the baseline methods to generate explanations for both node and graph instances.  For each experiment, we conduct $10$ times with random parameter initialization and report the average results as well as the standard deviation.  Detailed experimental setups are put in the appendix.

\subsection{Quantitative Evaluation}

\subsubsection{Comparison to Baselines.} 
We adopted the well-established experimental framework~\cite{ying2019gnnexplainer,luo2020parameterized,shan2021reinforcement}, where the explanation problem is framed as a binary classification of edges. Within this setup, edges situated inside motifs are regarded as positive edges, while all others are treated as negative. The importance weights offered by the explanation methods are treated as prediction scores. An effective explanation method, therefore, would assign higher weights to edges located within the ground truth motifs as opposed to those outside. To quantitatively evaluate the performance of these methods, we employed AUC as our metric. The average AUC scores and the associated standard deviations are reported in Table~\ref{tab:baselines}.  We observe that with a manually selected value for $k$, \ours consistently outperforms GNNExplainer and PGExplainer and competes with high-performing models like RGExplainer and DEGREE. The comparison demonstrates that our \ours considers locality, providing more accurate, comprehensive explanations and mitigating common locality pitfalls seen in other models.

\subsection{Model Analysis}
\label{sec:sim}

\subsubsection{Effectiveness of Bootstrapping Algorithm.} 
To directly show the effects of $k$ in \ours. We change the value of $k$ from 1 to 60 and show the resulting performance in Table~\ref{tab:tuningK}. We observe that, in general, a higher value of $k$ leads to improved performance. The reason is that large $k$ in \ours mitigates the locality and lossy aggregation losses in parametric explainers as discussed previously.  We use an underline to indicate the upper-bound for the value of $k$ suggested by the bootstrap algorithm in  Section~\ref{sec:met}. It should be noted that this upper-bound is particularly relevant to multi-motif scenarios considered in Theorem \ref{th:2}. 
Restricting to values of $k$ that are less than or equal to the suggested upper-bound achieves the best performance in the multi-motif MUTAG task, which is aligned with our theoretical analysis.

\subsubsection{Effects of Lossy Aggregation.} 
To evaluate the effects of lossy aggregation, we consider the  BA-Community and Tree-Cycles in this part.  As shown in Table~\ref{tab:tuningK}, \ours significantly outperforms PGExplainer. 
The reason is that the \ours partially mitigates the aggregation loss in GNN explanation methods by combining the outputs of multiple MLPs, hence combining multiple `weak' explainers into a stronger one.  In addition, we observe that
the performances of \ours are positively related to $k$. Next, we increase the dimensionality of hidden representation in the GNN model from 20 to 80. This reduces the loss in aggregation as at each layer several low dimensional vectors are mapped to high dimensional vectors. The explanation performances are shown in Table~\ref{tab:lossy}. For these two datasets, the performance improves as $k$ is increased when the dimension is 20, due to the mitigation of the aggregation loss, however, as expected, no improvement is observed when increasing  $k$ in explaining the GNN with dimension 80, since there is no significant aggregation loss to mitigate in that case.

\section{Conclusion}
In this work, we theoretically investigate the trivial solution problem in the popular objective function for explaining GNNs, which is largely overlooked by the existing post-hoc instance-level explanation approaches.  We point out that the trivial solution is caused by the signal problem and propose a new GIB objective with a theoretical guarantee. On top of that, we further investigate the locality and lossy aggregation issues in existing parametric explainers and show that most of them can be unified within the local explanation Methods, which are weak at handling real-world graphs, where the mapping between labels and motifs is one-to-many.  We propose a new factorization-based explanation model to address these issues.  Comprehensive experiments are conducted to verify the effectiveness of the proposed method.

\section{Acknowledgments}
This project was partially supported by NSF grants IIS-2331908 and CCF-2241057. The views and conclusions contained in this paper are those of the authors and should not be interpreted as representing any funding agencies.


\bibliography{aaai24}

\clearpage
\appendix

\section*{\centering \Large{Appendix}}

\section{Proofs}
\subsection{Proof of Lemma \ref{lem:1}}
\label{App:lem:1}
Note that from the data processing inequality and $B\leftrightarrow C \leftrightarrow A'$, we have:
\begin{align*}
    I(A',B)\leq I(A',C).
\end{align*}
Also, from $P_{A|C}=P_{A'|C}$, we have: 
\begin{align*}
    I(A,C)=I(A',C).
\end{align*}
Lastly, from $A\leftrightarrow B \leftrightarrow C$, we have:
\begin{align*}
    I(A,C)\leq I(A,B).
\end{align*}
Combining these three results, we get:
\begin{align*}
    I(A',B)\leq I(A',C)= I(A,C)\leq I( A,B).
\end{align*}
\qed

\subsection{Proof of Theorem \ref{th:1}}
\label{App:th:1}
Let us define $\gamma_{\alpha}\triangleq \min_{G'} {I(G, G')+\alpha H(Y|G')}$, and let $G^*_{\alpha}$ be an explanation achieving $\gamma_{\alpha}$.  Define $G'_{\alpha}$ as a random graph generated conditioned on $h(G)$ such that $P_{G'_{\alpha}|h(G)}=P_{G^*_{\alpha}|h(G)}$ and the Markov chain $G,G^*_{\alpha} \leftrightarrow h(G) \leftrightarrow G'_{\alpha}$ holds. It can be observed that by construction the conditions in Lemma \ref{lem:1} are satisfied for $A\triangleq G^*_{\alpha}, A'\triangleq G'_{\alpha}, B\triangleq G$ and $C \triangleq h(G)$. Hence, we have $I(G'_{\alpha},G)\leq I(G^*_{\alpha},G)$. As a result:
\begin{align*}
\gamma_\alpha=I(G,G^*_{\alpha})+\alpha H(Y|G^*_{\alpha}) &
\stackrel{(a)}{\geq}
I(G,G'_{\alpha})+\alpha H(Y|G^*_{\alpha})
\\& \stackrel{(b)}{=}
I(G,G'_{\alpha})+\alpha H(Y|G'_{\alpha})
\end{align*}
  where in (a) follows from $I(G'_{\alpha},G)\leq I(G^*_{\alpha},G)$ and (b) follows from the fact that 
the task is statistically degraded, the Markov chain $G^*_{\alpha}G'_{\alpha}\leftrightarrow h(G) \leftrightarrow Y$ and $P_{G^*_{\alpha}|h(G)}=P_{G'_{\alpha}|h(G)}$. Consequently, $G'_{\alpha}$ is also an optimal explanation in the GIB sense, i.e. achieves $\gamma_{\alpha}$. 
The proof is completed by defining the explanation function as $\Psi_\alpha(G)=G'_{\alpha}$.
\qed

\subsection{Proof of Theorem \ref{th:2}}
\label{App:th:2}
   \textbf{Part a)}: Note that the optimal Bayes classifier rule is given by $f^*(g)= \argmax_{y\in \{0,1\}} P(y|G=g)$. If $\nexists i\in [s]:g_i\subseteq g$, then $P(Y=1|G=g)=0$ and hence $f^*(g)=0$ as desired. So, it suffices to show that if $\exists i\in [s]:g_i\subseteq g$, then $P(Y=1|G=g)>P(Y=0|G=g)$. Let $\mathcal{I}=\{i\in [s]| g_{i} \subseteq g\}$.
   Note that
   \begin{align*}
        &P(Y=1|G=g)>P(Y=0|G=g) \iff \\& P(G=g,Y=1)>P(G=g,Y=0).
    \end{align*}
    Also,
    \begin{align}
        P(G=g,Y=1)&\stackrel{(a)}{\geq} \prod_{i\in \mathcal{I}}P(g_i\subseteq G_e)P(g-\cup_{i\in \mathcal{I}}g_i\subseteq G_0\subseteq g)
        \nonumber \\& 
        \stackrel{(b)}{=}  (\prod_{i\in \mathcal{I}}p_i)P(g-\cup_{i\in \mathcal{I}}g_i\subseteq G_0\subseteq g)
        \label{eq:th2:1}
    \end{align}
    where (a) follows from the facts that i) if all indicator motifs are equal to one then $Y=1$, ii) the indicator motifs are independent of each other, and iii) $G_e$ and $G_0$ are independent of each other, and (b) follows from the fact that the indicator motifs are Bernoulli variables with parameters $p_i, i\in [s]$.  On the other hand:
    \begin{align*}
        &P(G=g,Y=0)\stackrel{(a)}{\leq}  P(G_0=g) \\& \stackrel{(b)}{=} \prod_{i\in \mathcal{I}}P(g_i\subseteq G_0)P(g-\cup_{i\in \mathcal{I}}g_i= G_0-\cup_{i\in \mathcal{I}}g_i),
    \end{align*}
    where (a) follows from the fact that if $Y=1$ all indicator motifs are equal to zero, and 
in (b)  we have used the fact that the graph $G_0$ is Erd\"os-R\'enyi, and the motif subgraphs do not overlap. So, 
      \begin{align}
        P(G=g,Y=0)\leq  (\prod_{i\in \mathcal{I}}p_{i})P(g-\cup_{i\in \mathcal{I}}g_i\subseteq  G_0\subseteq g),
        \label{eq:th2:2}
    \end{align}
    where we have used the fact that 
    from condition iv) in the problem formulation, we have $P(g_i\subseteq G_0)\leq p_{i}, i\in [s]$. Combining \eqref{eq:th2:1} and \eqref{eq:th2:2} completes the proof of a).
    \\\textbf{Part b):} We provide an outline of the proof. It can be observed that as $\alpha\to \infty$, the optimal explanation algorithm in the modified GIB sense is the one that minimized the cross-entropy term $CE(Y,Y')$. We argue that the optimal explanation algorithm outputs one of the motifs present in $g$ if $Y=1$. The reasons is that in this case, $f^*(g)=f^*(\Psi(g))$ for all input graphs $g$, and the cross-entropy term is minimized by the assumption that $f^*$ is the optimal Bayes classifier rule. 
    
    Next, we argue that an r-local explanation method cannot produce the optimal explanation. The reason is that
    due to condition ii) and iii) in the problem formulation and definition of r-local explanation methods, the probability that an edge in a given motif is included in the explanation only depends on the presence of the corresponding motif in graph $g$ and not the presence of the other motifs. To see this, let $p_{j,j'}$ be the probability that the edge $(v_j,v_{j'})$ in motif $g_i\subseteq g$ is included in the explanation $\Psi(g)$. Then, by linearity of expectation, we have: 
    \[\mathbb{E}(|\Psi(G)|)\geq \sum_i p_i \sum_{j,j': (v_j,v_{j'})\text{edge in $g_i$}}p_{j,j'}.\] 
    So, we must have: 
    \[\sum_i p_i \sum_{j,j': (v_j,v_{j'})\text{edge in $g_i$}}p_{j,j'}\leq \gamma.\]  Consequently, the r-local explanation method cannot  output one of the motifs present in $g$ with probability one since the edge probabilities $p_{j,j'}$ cannot be equal to one as their summation should be less than $\gamma$. This completes the proof of part b).\\
    \textbf{Part c):} We construct an optimal explainer as follows. Let $k=s$ and define $\Psi_i(g)\triangleq g_i\mathbb{1}(g_i\in g)$. Furthermore, define the explainer $\Psi^*(g)\triangleq \argmin_{g_i, i\in [s]} \{i|g_i\subseteq g\}$. Then, since for $Y=1$, the output of $\Psi^*$ is a motif that is present in the input graph, as explained in the proof of Part b), $\Psi(g)$ is an optimal explanation function in the modified GIB sense as $\alpha\to \infty$ as it yields the minimum cross-entropy term. Furthermore, $\Psi(g)\in \{\Psi_1(g),\Psi_2(g),\cdots,\Psi_{k}(g)\}$ by construction. This completes the proof. 
\qed
\subsection{Proof of Proposition \ref{prop:2}}
\label{App:prop:2}
  Following the arguments in the proof of Theorem \ref{th:2}, it suffices to show that there exist  $\Psi_t, t\in [k^*]$ and $\Psi(\cdot)\in \{\Psi_1(\cdot),\Psi_2(\cdot),\cdots,\Psi_{k^*}(\cdot)\}$, such that $\Psi(g)$ is equal to a motif that is present in $g$ whenever $Y=1$. To construct such an explainer, let $\mathsf{P}=\{\mathcal{P}_t,t\in [k^*]\}$ be the minimal $k^*$-cover of $G_e$, and let $g_{i^*}$ be the motif in the input graph with the smallest index among all motifs present in the input graph. We take
    $\Psi_t, t\in [k^*]$ to be such that it assigns sampling probability one to the edges in $\mathcal{P}_t$ belonging to the motif $g_i$ in the input graph with the smallest index among all motifs whose edges overlap with  $\mathcal{P}_t$ and sampling probability zero to every other edge, i.e. it outputs the motif with the smallest index in its computation graph with probability one. Let $\Psi_0$ be such that it assigns probability one to an $\Psi_t, t\in [k^*]$ for which $\mathcal{P}_t$ overlaps with $g_{i^*}$ and zero to all other $\Psi_t$. Then, it is straightforward to see that the resulting sampled graph from K-FactExplainer $G^*$  would be equal to the motif with the smallest index among the motifs present in the input graph $g_{i^*}$. This completes the proof. 
    \qed
\section{Algorithm}
The detailed algorithm for Bootstrapping is shown in Alg.~\ref{alg:selectK}.

\begin{algorithm}[h]
  \caption{Bootstrapping Algorithm to Bound the Number of MLPs}
  \label{alg:selectK}
  \begin{algorithmic}[1] 
  \REQUIRE the GNN model to be explained, $f$, a set of graphs $\sG$, an explainer $\Psi$.
  \STATE $\hat{k} \leftarrow 0$. 
    \FOR{each graph $G=(\gV,\gE,\mathbf{X},\mathbf{A}) \in \sG$}
    \STATE $\mM \leftarrow \Psi(f,G)$ \# get a mask of edges by applying $g$ to explain $f$ on $G$. 
    \STATE $\vm \leftarrow \frac{\text{sum}(\mM,0)+\text{sum}(\mM,1)}{2}$ \# get all node in the explanation graphs. 
    \STATE $\sV_e \leftarrow \{v_i | \evm_i>0\}$ \# get all node in the explanation graphs. 
    \STATE Rank the nodes in $\sV_e$ according to their degrees/betweenness. 
    \STATE $k'\leftarrow 0$
    \WHILE{$\sV_e$ is not empty}
        \STATE get a node $v$ from $\sV_e$ with smallest degree;
        \STATE $G^{(l)}_v \leftarrow$ $l$-hop neighborhood graph with $v$.
        \STATE remove nodes of $G^{(l)}_v$ from $\sV_e$
        \STATE $k'\leftarrow k'+1$
    \ENDWHILE
    \STATE $\hat{k} \leftarrow \max{(k',\hat{k})}$.
    \ENDFOR
    \RETURN $\hat{k}$
  \end{algorithmic}
\end{algorithm}

\section{Detailed Experimental Setups}

\subsection{Experimental Setup}
In the experimental study, we first describe the synthetic
and real-world datasets used for experiments,  baseline methods, and experimental setup. After qualitative and quantitative analysis, we demonstrate that the K-FactExplainer is effective for explaining both node and graph instances, and it outperforms state-of-the-art explanation methods in various scenarios. We attribute the gains of K-FactExplainer to the mitigation of two types of performance losses, which were studied analytically in prior sections, namely, lossy aggregation and locality. 

As mentioned in Remark \ref{rem:loss}, the first type of loss, aggregation loss, is due to the fact that the aggregator function at each layer of the GNN is not one-to-one, and hence loses information. This is particularly true if the mid-layers of the GNN have similar (low) dimensions. The second type of loss, the locality loss, manifests in multi-motif scenarios as shown in Theorem \ref{th:2}. In order to study each of these types of losses in isolation, we first focus on single-motif tasks and study lossy aggregation, then, we focus on multi-motif tasks and study the locality loss. 

To prove that the explainer's performance gains in single-motif tasks are indeed due to the mitigation of aggregation loss, it suffices to show that if the lossy aggregator is replaced by a (almost) lossless aggregator, the performance gains vanish. One can construct a GNN with almost lossless aggregation by choosing the dimensionality of the layers of the GNN to be sequentially increasing. 
Hence, to demonstrate our claim, we consider several single-motif classification tasks and show that increasing the hyperparameter $k$ leads to performance gains if the GNN has similar low-dimensional mid-layers, whereas if the layers increase in dimension sequentially, no performance gain is observed when increasing $k$, thus confirming the hypothesis that the K-FactExplainer gains in single-motif scenarios are indeed due to lossy aggregation. To demonstrate the fact that the K-FactExplainer mitigates the locality loss suffered by conventional local explainers, we consider several multi-motif classification tasks, and show that the explainer's accuracy increases as $k$ is increased in accordance with the analytical results of Theorem \ref{th:2} and Algorithm \ref{alg:selectK}. 

\subsubsection{Datasets.} 
Six benchmark datasets are used for evaluation, with \bashapes, \bacom, \treec, and \treeg~\cite{ying2019gnnexplainer} for the node classification task, and \bamo~\cite{luo2020parameterized} and
MUTAG~\cite{debnath1991structure} for the graph classification task. \bashapes is a dataset generated using the Barabási-Albert (BA) graph randomly attached with ``house''-structured network motifs. \bacom, also generated using the BA model, focuses on community structures within the graph, with nodes connected based on a preferential attachment model. \treec is a dataset where cycle structures are embedded within trees, created by introducing cycles by connecting nodes in the trees. Lastly, \treeg is a dataset that combines tree and grid structures by embedding grid structures within trees. {\mutag} is a real-world dataset that comprises chemical compound graphs, where each graph represents a chemical compound, and the task is to predict whether the compound is mutagenic or not. These synthetic and real datasets allow us to evaluate the performance of the \ours and baseline methods in controlled settings, providing insights into their effectiveness and interpretability for node classification tasks.

\subsubsection{Baselines.} 
To assess the effectiveness of the proposed framework, we compare our method with representative GIB-based explanation methods, GNNExplainer~\cite{ying2019gnnexplainer} and PGExplainer~\cite{luo2020parameterized}. We also include other types of explanation methods, RG-Explainer~\cite{shan2021reinforcement} and DEGREE ~\cite{feng2022degree}, GRAD~\cite{ying2019gnnexplainer}, and ATT~\cite{veličković2018graph} for comparison.  Specifically,
GRAD computes gradients of GNN’s objective function w.r.t. each edge for its importance score. (2) ATT averages the edge attention weights across all layers to compute the edge weights. (3) GNNExplainer~\cite{ying2019gnnexplainer} is a post-hoc method, which provides explanations for every single instance by learning an edge mask for the edges in the graph. The weight of the edge could be treated as important. (4) PGExplainer~\cite{luo2020parameterized} extends GNNExplainer by adopting a deep neural network to parameterize the generation process of explanations, which enables PGExplainer to explain the graphs in a global view. It also generates the substructure graph explanation with the edge importance mask. (5) RGExplainer~\cite{shan2021reinforcement} is an RL-enhanced explainer for GNN, which constructs the explanation subgraph by starting from a seed and sequentially adding nodes with an RL agent. (6) DEGREE ~\cite{feng2022degree} is a decomposition-based approach that monitors the feedforward propagation process in order to trace the contributions of specific elements from the input graph to the final prediction.

For each dataset, we train a graph neural network model to perform the node or graph classification task. Each model is a three-layer GNN with a hidden size of 20, followed by an MLP that maps these embeddings to the number of classes. After training the model, we apply the \ours and the baseline methods to generate explanations for both node and graph instances.  For each experiment, we conduct $10$ times with random parameter initialization and report the average results as well as the standard deviation. 

\section{Extra Experiments}

\subsection{Model Analysis}
\label{App:sec:sim}

\textbf{Effects of Locality in Multi-motif Tasks.}
So far, we have illustrated the gains due to the mitigation of the locality loss by considering the multi-motif MUTAG classification task (Table \ref{tab:baselines}). To further empirically demonstrate these gains, in this section,  we introduce a new graph classification dataset, {\bafour} which includes 1000 stochastically generated graphs. The generation process follows the procedure described in \eqref{eq:2}. That is, for each graph, we first generate a base graph $g_0$ according to the BA model. Then, the graph is assigned a binary label randomly and uniformly (by generating $E^s$, the indicators of different motifs). Each label is associated with two motifs. The motif graph ($g_e$ in \eqref{eq:2}) is generated based on $E^s$, and finally the graph $g$ is produced by taking the union of $g_0$ and $g_e$.  

As shown in Theorem \ref{th:2}, assuming that the classifier to be explained has optimal performance, the optimal explainer (in the modified GIB sense) outputs one of the motifs that are present in the graph. Consequently, we define the \textit{motif coverage rate} (CR) as a measure of the performance of the explainer as follows. Recall that the explainer assigns a probability of being part of the explanation to each of the graph edges, e.g., the probability vector $\overline{\Omega}$ in Figure \ref{fig:Ov}. Let $\mathcal{E}_r$ denote the $r$ top-ranked edges in terms of probability of being part of the explanation,  where $r$ is the maximum motif size. 
For each motif $g_i\in g$ with edge set $\mathcal{E}_i$, we define its coverage rate as $CR_i\triangleq \frac{|\mathcal{E}_i\cap \mathcal{E}_r|}{\max_i |\mathcal{E}_i|}$, i.e., the fraction of the motif's edges that are top-ranked. The CR is defined as $\max_i CR_i$, the maximum of coverage rate among all motifs. Table ~\ref{tab:batest} shows that with more complex datasets, our method will significantly improve the faithfulness of the explanation in terms of coverage rate.

\begin{figure}
  \centering
  \subfigure[Base BA-graph]{\includegraphics[width=1.1in,,draft=false]{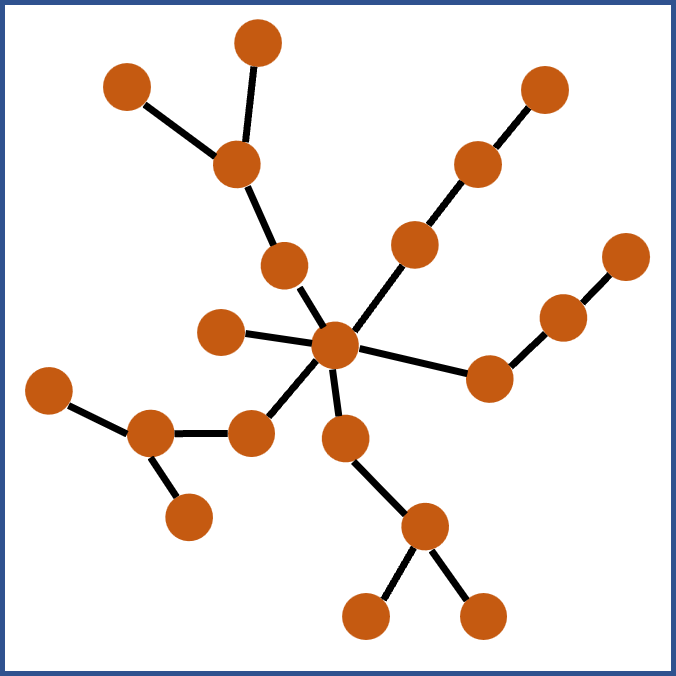}\label{fig:app:base}}\hspace{-0.2em}
  \subfigure[Motifs in Label 0]{\includegraphics[width=1.1in,,draft=false]{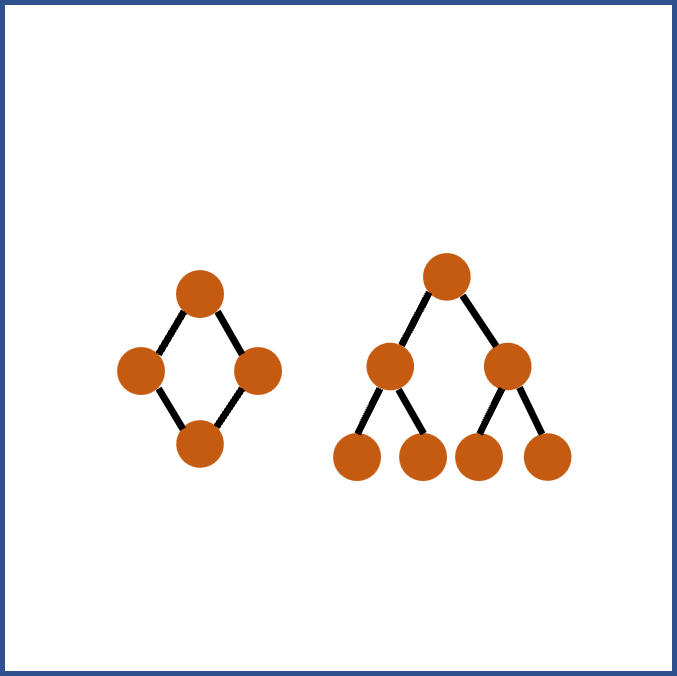}\label{fig:::label0}}\hspace{-0.2em}
\subfigure[Motifs in Label 1]{\includegraphics[width=1.1in,,draft=false]{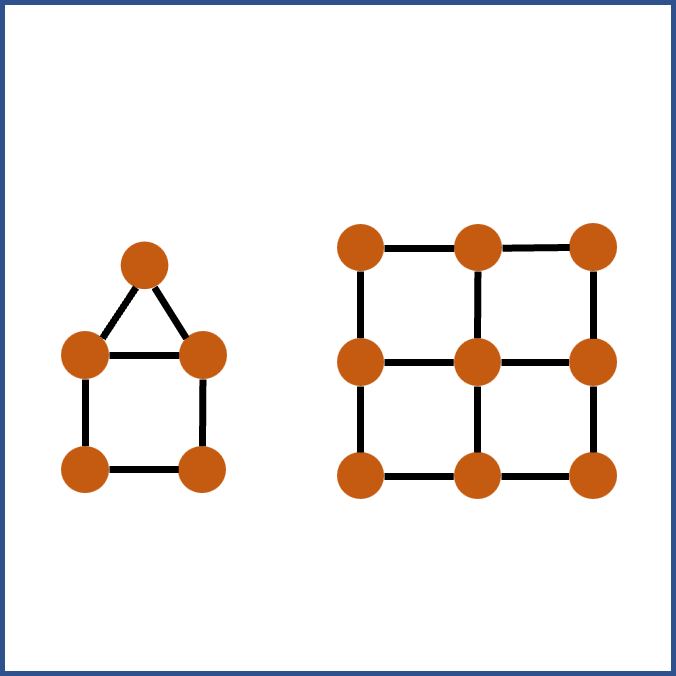}\label{fig:::label1}}
  \caption{Visualization of {\bafour} dataset. A BA-graph is used as the base graph. Each label is associated with two motifs.}
 \label{fig:ba4}
\end{figure}

\begin{table*}[h!]
\centering
  \caption{Average Coverage Rate for \bafour.} 
    \label{tab:batest} 
\scalebox{0.9}{
  \begin{tabular}{c|ccccc}
    \hline
 & GNNExp. & PGExp. & k=1&  k=2& k=3 \\
    \hline
    Coverage Rate (CR) & 0.114& 0.433 & 0.442& 0.615& \textbf{0.712}\\
    AUC & 0.737& 0.755& 0.761& 0.772& \textbf{0.789}\\
    \hline
  \end{tabular}
}
\end{table*}

\subsection{Efficiency Analysis} The proposed method comprises networks capable of generating inductive explanations for new instances across a population. Following~\cite{luo2020parameterized}, we measure the inference time, which is the time needed to explain a new instance with a trained model. Table~\ref{tab:time} presents the running time for \ours with various $K$ values. The results indicate that the inference time of \ours is comparably similar to that of the PGExplainer, which is one of the most sufficient explanation techniques. 

\begin{table*}[ht]
\centering
\scalebox{0.9}{
  \begin{tabular}{c|cccccc}
    \hline
 & \bashapes & \bacom & \treec& \treeg &\bamo &\mutag \\
    \hline
PGExp. & $4.27$& $6.48$& $0.50$& $0.55$& $0.44$& $2.48$\\
K=1 & $4.18$& $6.16$& $0.46$& $0.53$& $0.44$& $3.01$\\
K=2 & $4.26$& $6.26$& $0.49$& $0.57$& $0.47$& $2.50$\\
K=3 & $4.47$& $6.88$& $0.55$& $0.62$& $0.55$& $2.86$\\
K=5 &$4.27$& $6.29$& $0.53$& $0.58$& $0.52$& $2.56$\\
K=10 & $4.39$& $6.42$& $0.58$& $0.69$& $0.53$& $2.78$\\
K=20 & $4.69$& $6.81$& $0.66$& $0.85$& $0.65$& $3.17$\\
K=60 & $5.75$& $8.39$& $1.38$& $1.20$& $1.14$& $5.79$\\
\hline
  \end{tabular}
  }
  \caption{K-Explainer Inference Time (ms)}
    \label{tab:time} 
\end{table*}

\subsection{Qualitative evaluation}
We selected an instance to visually demonstrate the explanations given by  GNNExplainer, PGExplainer, and {\ours} in Table ~\ref{tab:qualitative}. 

\begin{table*}[h!]
  \centering
  \resizebox{\textwidth}{!}{%
  \begin{tabular}{  c  c  c  c  c | c  c  c }
    &\multicolumn{4}{c}{\textbf{Node Classification}}&\multicolumn{3}{c}{\textbf{Graph Classification}}\\
     & BA-Shapes & BA-Community & Tree-Cycles & Tree-Grid & BA-2Motifs & MUTAG& BA-4Motifs  \\ 
        {GNNExplainer}    
    &\raisebox{-.5\height}{
    \includegraphics[width=0.2\linewidth, height=0.15\linewidth]{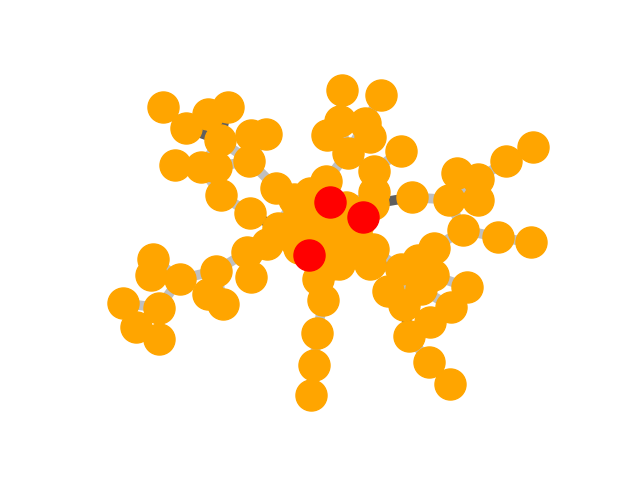}}
    &\raisebox{-.5\height}{
    \includegraphics[width=0.2\linewidth, height=0.15\linewidth]{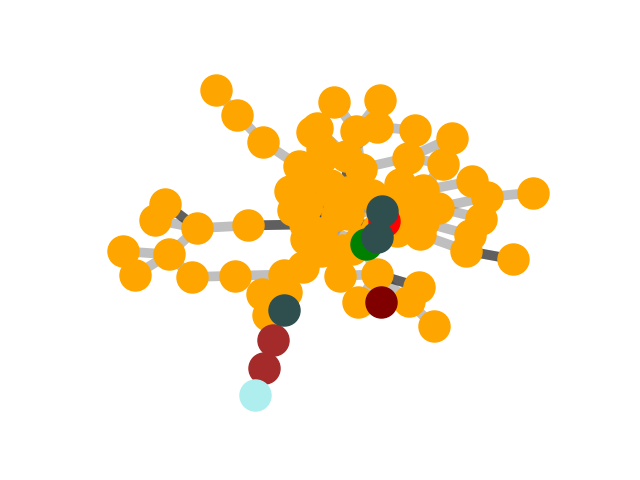}}
    &\raisebox{-.5\height}{
    \includegraphics[width=0.2\linewidth, height=0.15\linewidth]{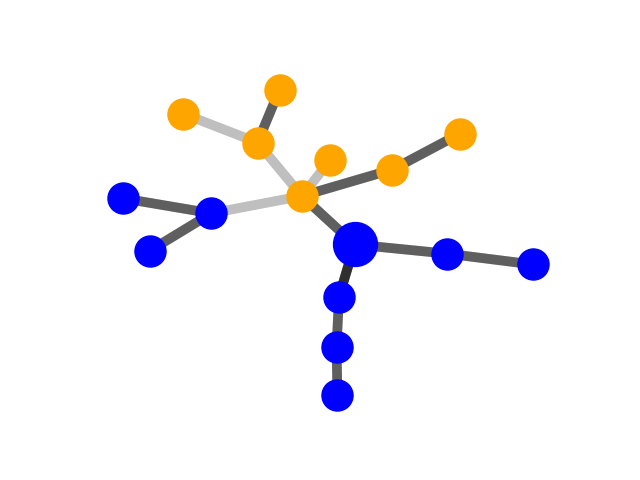}}
    &\raisebox{-.5\height}{
    \includegraphics[width=0.2\linewidth, height=0.15\linewidth]{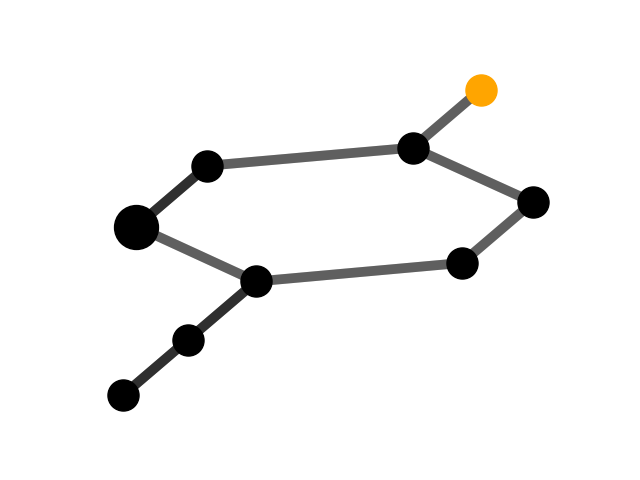}}
    &\raisebox{-.5\height}{
    \includegraphics[width=0.2\linewidth, height=0.15\linewidth]{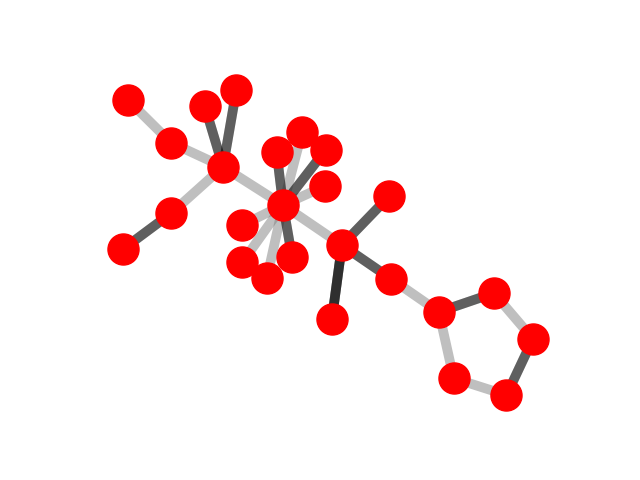}}
    &\raisebox{-.5\height}{
    \includegraphics[width=0.2\linewidth, height=0.15\linewidth]{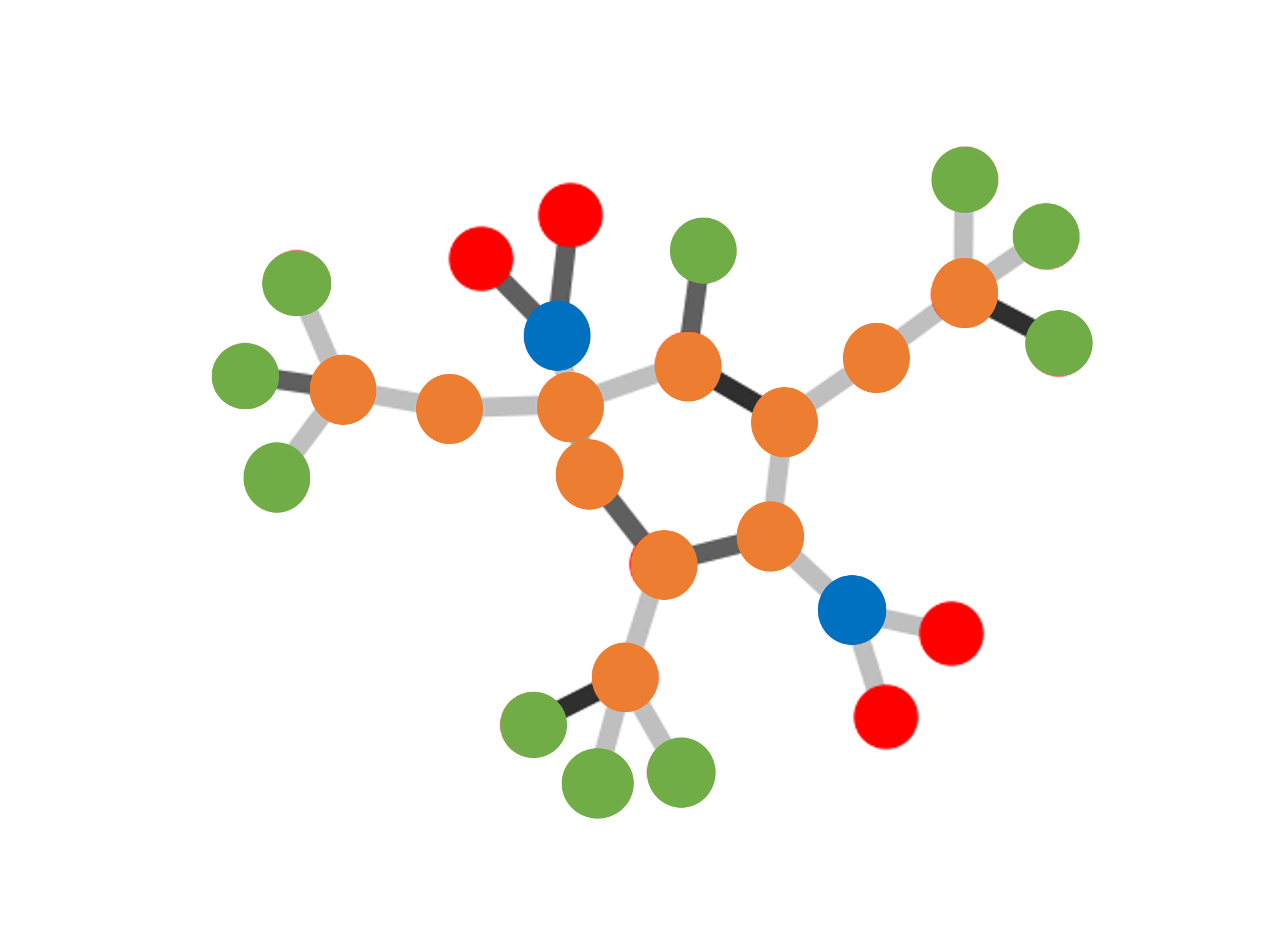}}
    &\raisebox{-.5\height}{
     \includegraphics[width=0.2\linewidth, height=0.15\linewidth]{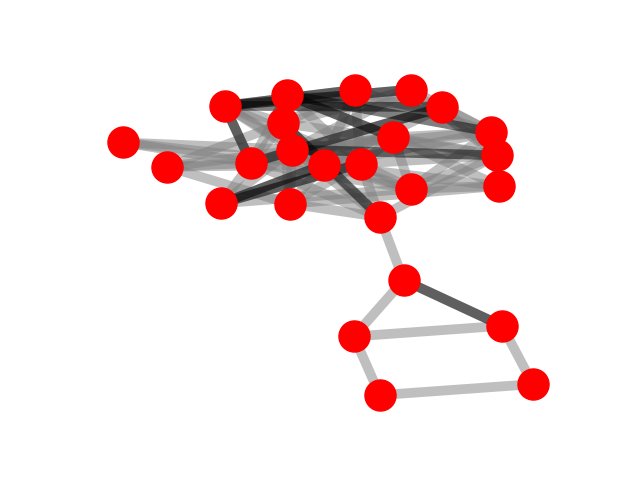}}\\
     
   \centering PGExplainer
    &\raisebox{-.5\height}{
    \includegraphics[width=0.2\linewidth, height=0.15\linewidth]{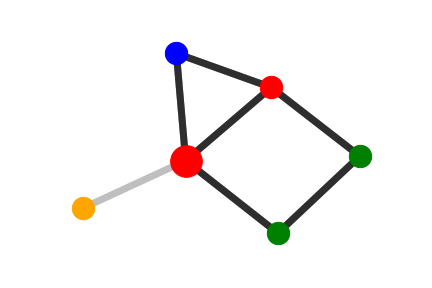}}
    &\raisebox{-.5\height}{
    \includegraphics[width=0.2\linewidth, height=0.15\linewidth]{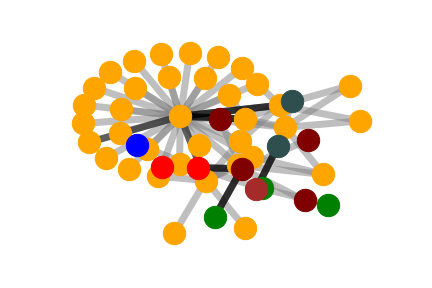}}
    &\raisebox{-.5\height}{
    \includegraphics[width=0.2\linewidth, height=0.15\linewidth]{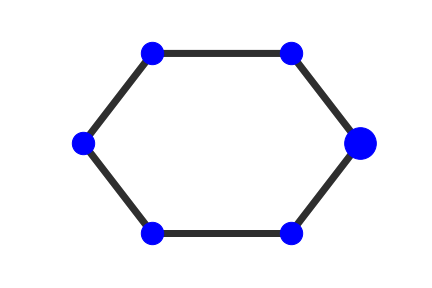}}
    &\raisebox{-.5\height}{
    \includegraphics[width=0.2\linewidth, height=0.15\linewidth]{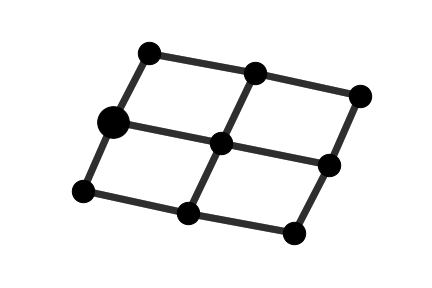}}
    &\raisebox{-.5\height}{
    \includegraphics[width=0.2\linewidth, height=0.15\linewidth]{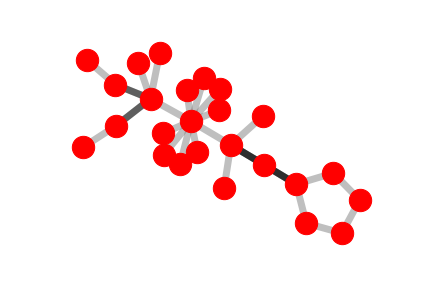}}
    &\raisebox{-.5\height}{
    \includegraphics[width=0.2\linewidth, height=0.15\linewidth]{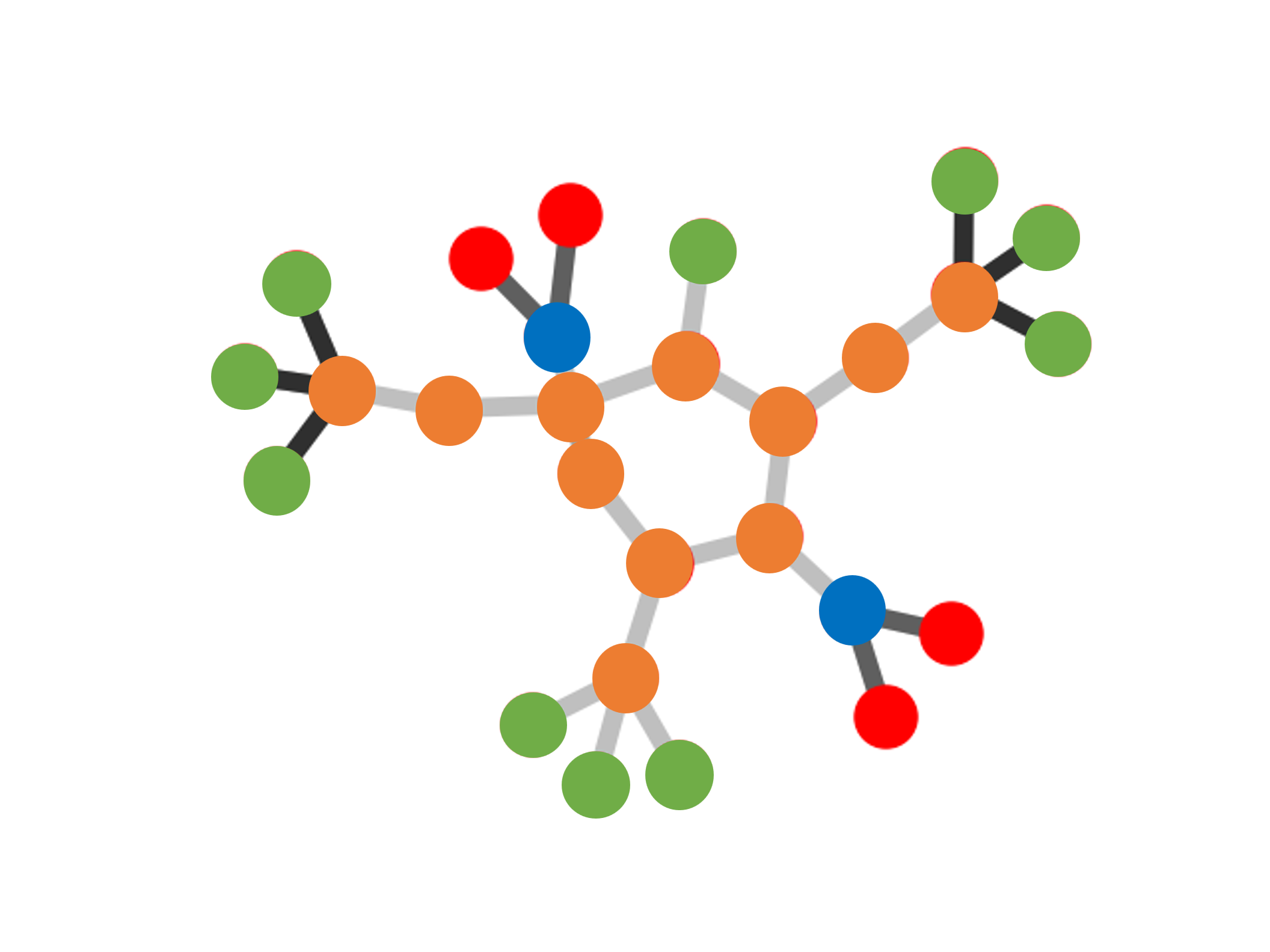}}
    &\raisebox{-.5\height}{
     \includegraphics[width=0.2\linewidth, height=0.15\linewidth]{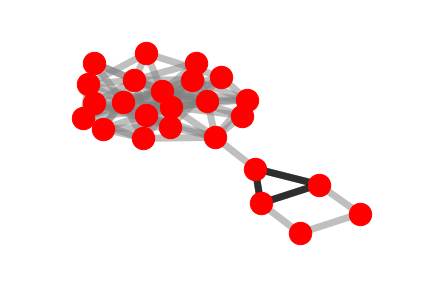}}\\
     
    \centering  K-FactExplainer
    &\raisebox{-.5\height}{
    \includegraphics[width=0.2\linewidth, height=0.15\linewidth]{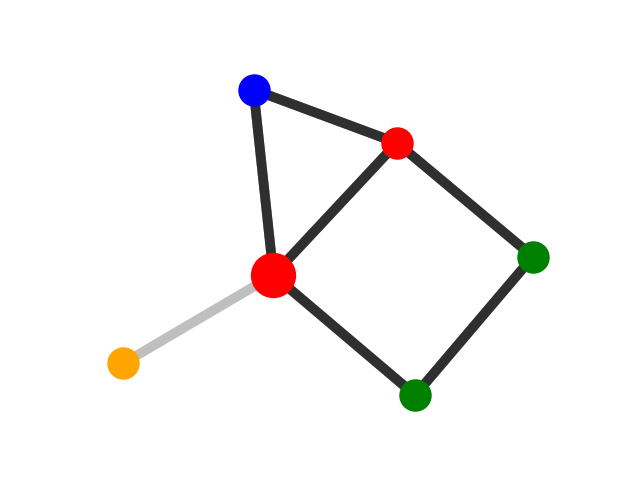}}
    &\raisebox{-.5\height}{
    \includegraphics[width=0.2\linewidth, height=0.15\linewidth]{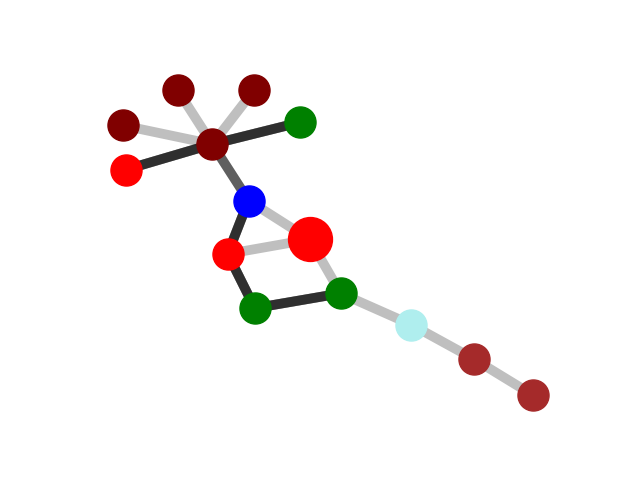}}
    &\raisebox{-.5\height}{
    \includegraphics[width=0.2\linewidth, height=0.15\linewidth]{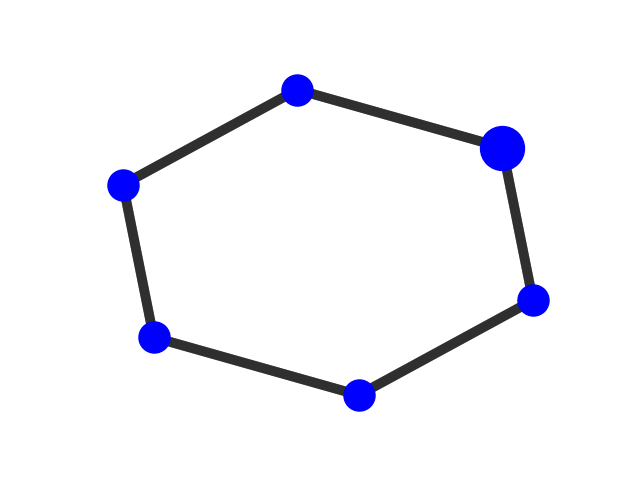}}
    &\raisebox{-.5\height}{
    \includegraphics[width=0.2\linewidth, height=0.15\linewidth]{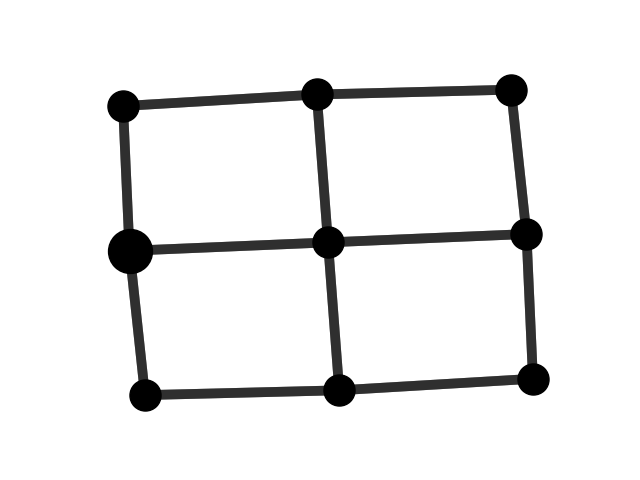}}
    &\raisebox{-.5\height}{
    \includegraphics[width=0.2\linewidth, height=0.15\linewidth]{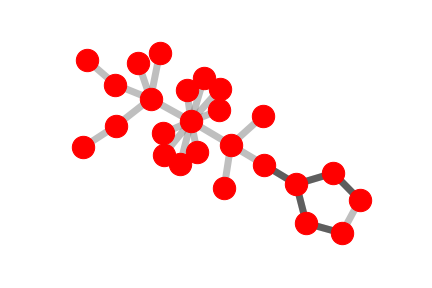}}
    &\raisebox{-.5\height}{
    \includegraphics[width=0.2\linewidth, height=0.15\linewidth]{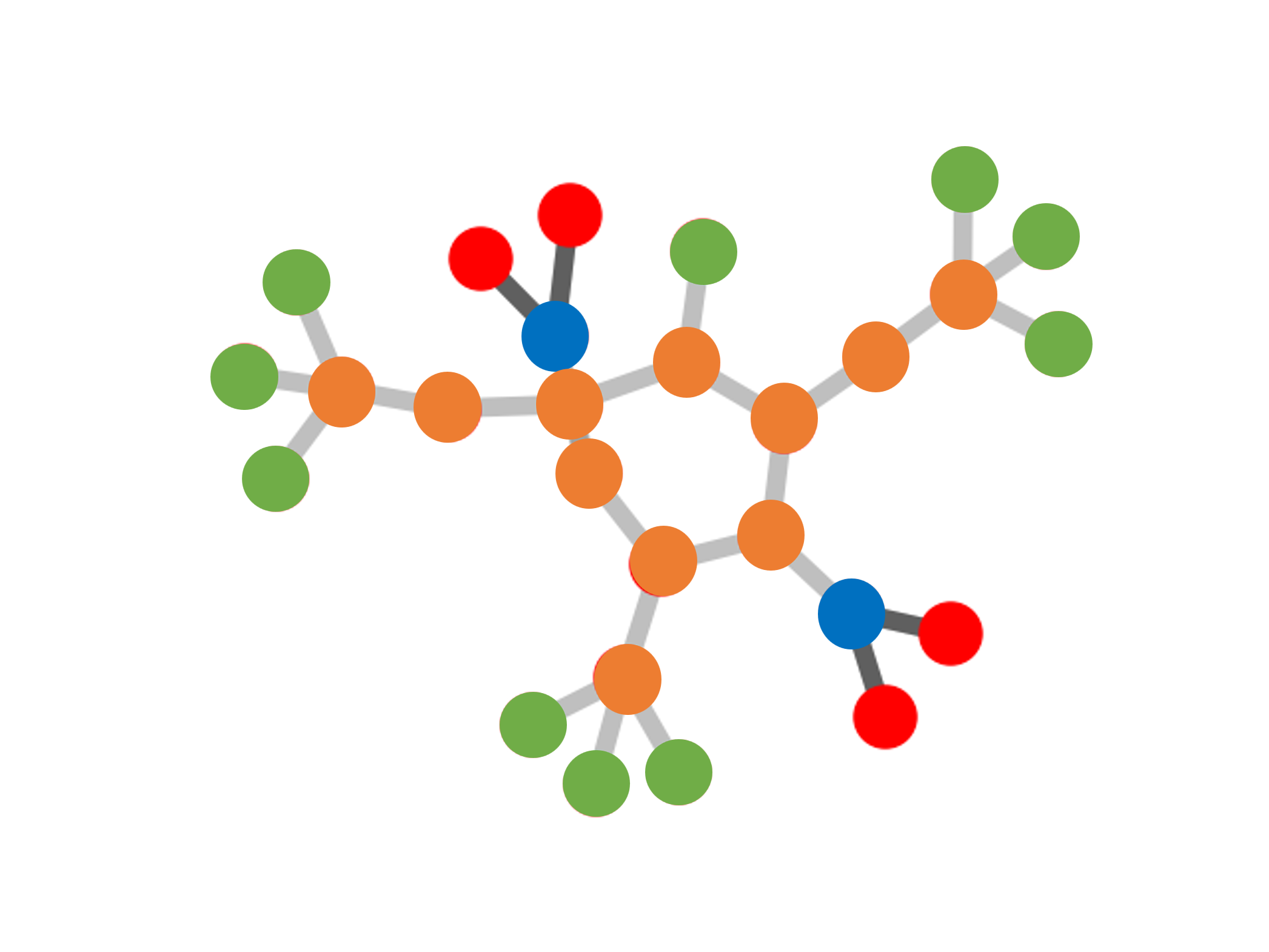}}
    &\raisebox{-.5\height}{
     \includegraphics[width=0.2\linewidth, height=0.15\linewidth]{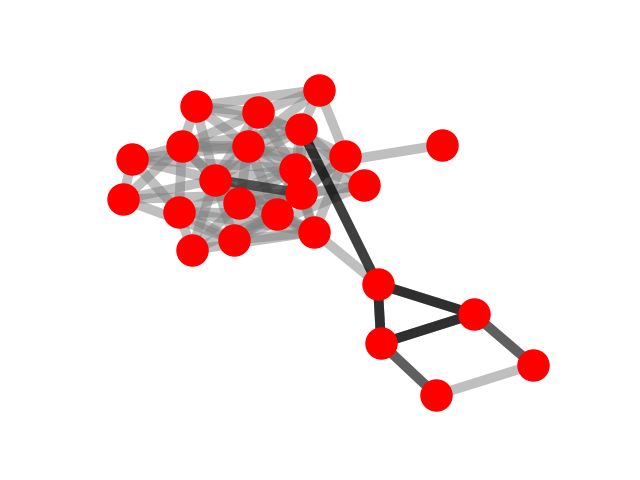}}
     \\ 
     \centering  Motifs
    &\raisebox{-.5\height}{
    \includegraphics[width=0.075\linewidth, height=0.1\linewidth]{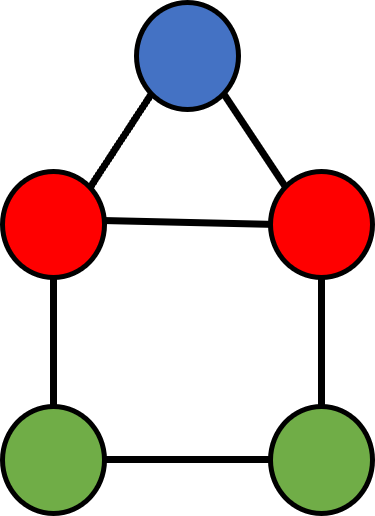}}
    &\raisebox{-.5\height}{
    \includegraphics[width=0.075\linewidth, height=0.1\linewidth]{figures/visual/house.png}}
    &\raisebox{-.5\height}{
    \includegraphics[width=0.1\linewidth, height=0.1\linewidth]{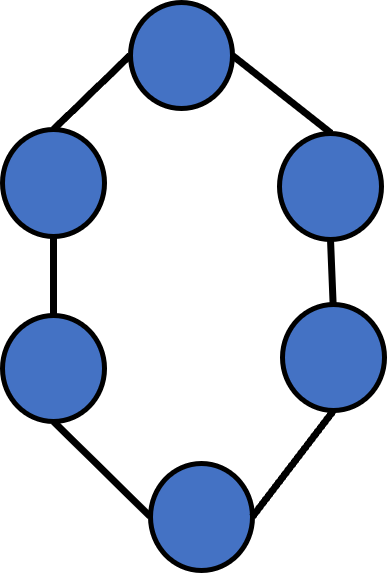}}
    &\raisebox{-.5\height}{
    \includegraphics[width=0.1\linewidth, height=0.1\linewidth]{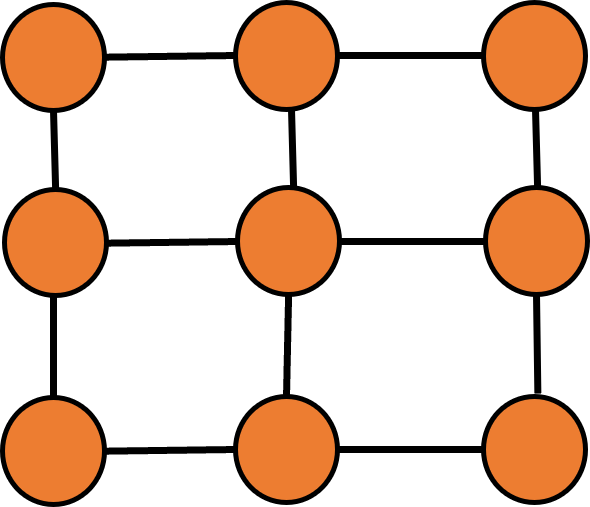}}
    &\raisebox{-.5\height}{
    \includegraphics[width=0.075\linewidth, height=0.1\linewidth]{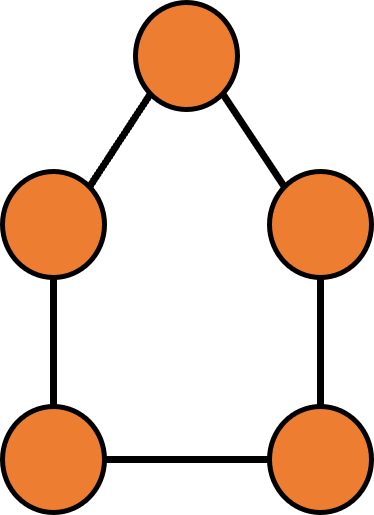}}
    &\raisebox{-.5\height}{
    \includegraphics[width=0.05\linewidth, height=0.05\linewidth]{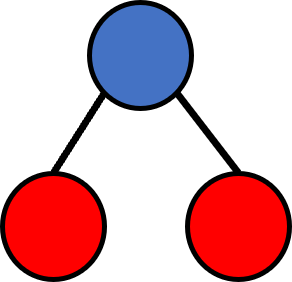}}
    &\raisebox{-.5\height}{
     \includegraphics[height=0.1\linewidth,draft=false]{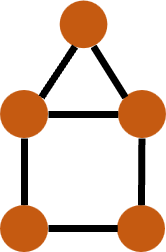}}
  \end{tabular}
  }
  \caption{Qualitative Analysis}
  \label{tab:qualitative}
\end{table*}

\end{document}